\documentclass{article} 
\usepackage{times}
\usepackage[margin=1in]{geometry}
\usepackage{natbib}

\usepackage{amsmath,amsfonts,bm}

\def\eqref#1{equation~\ref{#1}}

\def\1{\bm{1}}

\DeclareMathAlphabet{\mathsfit}{\encodingdefault}{\sfdefault}{m}{sl}
\SetMathAlphabet{\mathsfit}{bold}{\encodingdefault}{\sfdefault}{bx}{n}

\usepackage[hidelinks]{hyperref}
\usepackage{url}

\usepackage{booktabs}
\usepackage{amsfonts}
\usepackage{amsmath}
\usepackage{amsthm}
\usepackage{nicefrac}
\usepackage{microtype}

\usepackage{xcolor}
\usepackage{colortbl}
\usepackage[textsize=tiny]{todonotes}
\usepackage{amssymb}
\usepackage{mathtools}

\usepackage{cleveref}
\usepackage{algorithm}
\usepackage{algorithmic}
\usepackage{dsfont}

\usepackage{multirow}
\usepackage{graphicx}
\usepackage{makecell}
\usepackage{wrapfig}
\usepackage{tabularx}

\theoremstyle{plain}

\theoremstyle{definition}

\theoremstyle{remark}

\title{
Long-Horizon Textual World Modeling \\ through Structured Reasoning
}

\author{
Fangxin Wang$^{1,*}$ \quad
Xiang Gao$^{2,*}$ \quad
Yuguang Yao$^{2}$ \quad
\\
Kaiwen Dong$^{2}$ \quad
Nikash Walia$^{2}$ \quad
Kamalika Das$^{2}$\\[0.5em]
$^{1}$University of Illinois Chicago \quad
$^{2}$Intuit AI Research\\[0.25em]
\small \texttt{fwang51@uic.edu} \quad
\texttt{xiang\_gao@intuit.com} \quad
\texttt{kamalika\_das@intuit.com}\\[0.25em]
\small $^{*}$Equal contribution.
}

\date{}

\begin{document}

\maketitle

\begin{abstract}
World models must predict how an environment evolves under sequences of actions, enabling agents to compare possible futures and reason about counterfactual actions before acting. Long-horizon prediction is commonly obtained by recursively applying a one-step transition model, but intermediate errors can compound over time. Multi-step dynamics models instead condition on a sequence of future actions and predict their consequences directly, but become harder to learn as horizon grows: the model must track interacting state changes across the trajectory, endpoint supervision provides weak credit assignment, and intermediate predictions can remain plausible while losing information needed for later states. We show that these challenges can be addressed by casting the internal evolution of a multi-step transition as structured reasoning over textual world states: reasoning over sparse state changes reduces the burden of state tracking, a predictive-gain objective rewards the learned state for improving over a matched predictor that conditions on raw history instead, and intermediate predictive rewards supervise each state along the trajectory. Because these intermediate states are explicit textual representations of the world, they provide semantically meaningful targets that can be inspected, scored, and corrected during training. Across \textsc{ScienceWorld}, \textsc{Jericho}, and \textsc{CEO-Bench}, our approach achieves the strongest average long-horizon performance against recursive and non-recursive baselines that condition directly on raw history, with gains increasing at longer horizons: at horizon 10, it exceeds the strongest such baseline by 12.3\% relative on \textsc{ScienceWorld} and 9.3\% relative on \textsc{CEO-Bench}, and error on \textsc{Jericho} drops by nearly 30\%. In a controlled counterfactual study, our model is also the only one with statistically significant sensitivity to future actions ($p<0.001$), while baselines show null or anti-correlated sensitivity once ground-truth geometry is accounted for. These results suggest that making a multi-step transition's internal trajectory explicit and supervisable through structured reasoning is a better abstraction for long-horizon textual world modeling than recursively composing one-step predictions.

\end{abstract}

\section{Introduction}

World models enable agents to reason about the consequences of their actions before interacting with an environment. For language-based agents, this requires predicting how an environment will evolve under a sequence of actions from observations that reveal only part of the underlying world state \citep{li2026bridging,gu2025webdreamer,li2026word,zuo2026qwen}. A common approach is to learn a one-step transition model and recursively roll it forward, feeding each predicted state back into the model to produce the next \citep{hafner2019learning,hafner2020dream}. This factorization is convenient and naturally exposes intermediate states, but it also creates a fundamental weakness: errors in an intermediate prediction become part of the context for every subsequent transition and can therefore compound with horizon \citep{asadi2019combating,lambert2021learning}.

Non-recursive prediction offers an alternative. Rather than repeatedly applying a learned one-step transition to its own predictions, prior work directly predicts a future state at a target horizon from the current state and future action sequence \citep{asadi2019combating,lambert2021learning}. The difficulty is that long-horizon prediction is inherently multi-horizon: training a single direct model to accurately cover many horizons is challenging, while training separate models for different horizons fragments the data and prevents shared learning across them. These limitations are especially consequential in partially observable environments, where the model must preserve information from history while tracking action-dependent changes over many steps.

\begin{figure}[t]
\centering
\includegraphics[width=\linewidth]{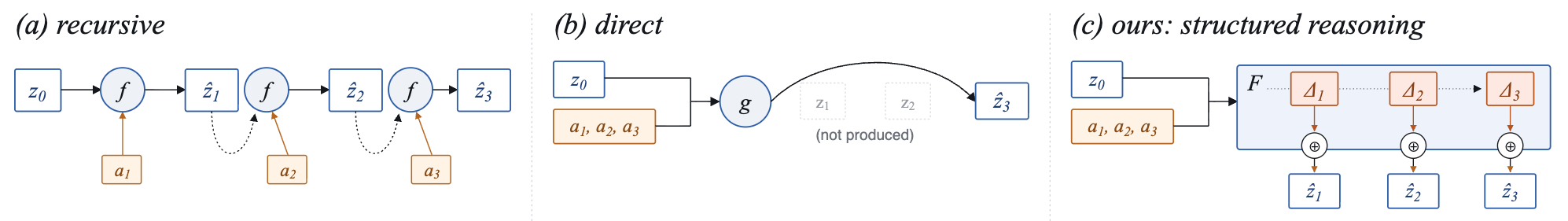}
\vspace{-1.5em}
\caption{\textbf{(a) Recursive} rollout feeds each predicted state back as input to the next call. \textbf{(b) Direct} prediction conditions on the entire action sequence and jumps straight to the endpoint, without exposing intermediate structure. \textbf{(c) Ours: structured reasoning} instead emits sparse deltas $\Delta_1,\ldots,\Delta_k$ in one autoregressive decode, applied ($\oplus$) to the persistent state to obtain $\hat z_1,\ldots,\hat z_k$ without re-entering the model: this eases state tracking and lets intermediate rewards localize errors.}
\label{fig:intro-teaser}
\end{figure}

The challenge, then, is to retain the benefits of non-recursive prediction while recovering the shared structure across horizons that direct endpoint prediction tends to discard. Rather than treating each horizon as a separate prediction target, we model the entire multi-step evolution as a single structured trajectory. Concretely, we cast this trajectory as reasoning over textual world states, where each action induces a sparse change to an otherwise persistent state (Figure~\ref{fig:intro-teaser}). This representation allows the same transition to support predictions at multiple horizons while making the internal evolution explicit. In this sense, direct endpoint prediction and our approach parallel the distinction between an LLM answering immediately and an LLM reasoning through a chain of thought before answering \citep{wei2022chain}: rather than free-form scratch text, however, our intermediate reasoning takes the form of an explicit, semantically grounded world-state trajectory that can itself be supervised and rewarded.

This structure also addresses the main difficulties of learning long-horizon dynamics jointly. Reasoning over sparse state changes reduces the burden of tracking interacting updates across a long action sequence while preserving unaffected facts. Exposing intermediate states provides semantically meaningful targets for supervision and process rewards, allowing errors to be localized throughout the trajectory rather than only at its endpoint. Finally, optimizing the predictive gain of the learned state over a matched predictor that conditions on raw history instead encourages it to retain information that may only become useful for downstream observations. Together, these mechanisms allow a single multi-step transition to share learning across horizons while avoiding recursive state feedback. Although the underlying language model remains autoregressive over tokens, predicted intermediate world states are not recursively fed back into separate transition calls.

We evaluate our approach on three partially observable interactive domains with complementary long-horizon dynamics: \textsc{ScienceWorld}~\citep{wang2022scienceworld}, \textsc{Jericho}~\citep{hausknecht19}, and \textsc{CEO-Bench}~\citep{chen2026ceo}. latent-SR achieves the strongest average long-horizon prediction performance across all three, with its advantage concentrated at longer horizons, and is also the only evaluated model with statistically significant action-conditioned sensitivity in a controlled counterfactual study, while baselines are null or significantly negative after correcting for ground-truth branch geometry. Ablations and controlled perturbations further separate the effects of transition factorization, training, and intermediate supervision, supporting a trajectory-explicit, non-recursive transition as a practical alternative to recursively composing one-step predictions.

\section{Related Work}
\label{sec: related work}

\textbf{Predictive states under partial observability.}
When the current observation does not fully characterize the environment, an agent must summarize its history into a state sufficient for future prediction. In POMDPs, this role is formalized by the belief state \citep{kaelbling1998planning}, while predictive state representations define state directly through action-conditional predictions of future observations \citep{littman2001predictive,singh2004predictive}. Learned recurrent and latent state-space models integrate past observations and actions into representations that mediate future dynamics \citep{downey2017predictive,hafner2019learning}, and information-state formulations characterize when such a representation can recursively replace the full history \citep{subramanian2022approximate}. The prediction horizon also affects what these representations retain: multi-future supervision distinguishes states with identical immediate outcomes \citep{dedieu2024learning}, and future-observation prediction improves history representations for memory and reasoning \citep{kwon2024empirical}. These works motivate our use of long-horizon prediction as supervision for the latent state, not merely as an evaluation target.

\textbf{Long-horizon dynamics in latent world models.}
Many world models for visual and control domains encode observations into fixed-dimensional continuous latents and learn a shared one-step transition, obtaining longer predictions by recursively applying it \citep{ha2018world}. This reuses the same dynamics model across horizons and exposes intermediate states, but repeated rollout can amplify local errors. The Dreamer line of latent-imagination agents \citep{hafner2019learning,hafner2020dream,hafner2021mastering,hafner2023mastering} and value-equivalent planning with a learned model \citep{schrittwieser2020mastering} are prominent instances at scale. Multi-step rollout objectives address error accumulation while retaining recursive dynamics: latent overshooting supervises predictions across multiple rollout distances \citep{hafner2019learning}, self-predictive representations recursively predict future latents \citep{schwarzer2021data}, and joint-embedding predictive models \citep{lecun2022jepa,assran2023ijepa,bardes2024vjepa} train on their own intermediate predictions \citep{terver2026what}. A complementary line instead changes the temporal factorization: direct multi-step models predict a farther-future state from the current state and action sequence \citep{asadi2019combating,lambert2021learning}, with related work supporting variable horizons, direct endpoint prediction, or action-level temporal abstraction \citep{du2026beyondnext,li2026beyondmyopic,park2026scalable}. These establish direct prediction as an alternative to deep recursive rollout, but target continuous latent spaces and predict only the endpoint state; our transition instead exposes an action-aligned intermediate state trajectory within the same non-recursive call, making the trajectory itself available for supervision.

\textbf{Text world models and textual latent states.}
Text world models use language models to predict action-conditioned outcomes from an interaction context or state \citep{li2026bridging, gao2026textual}, with recent work on web interaction, next-state prediction, and heterogeneous agent environments \citep{gu2025webdreamer,li2026word,zuo2026qwen}. Unlike continuous latents, textual states share their modality with the history, actions, and predicted observations. To ensure the learned state carries the information needed for prediction, our framework conditions prediction on the history only through this state. Generating a textual state requires the model to explicitly select and restate its contents, so important facts may be omitted during encoding or regeneration. The sequential structure of text also lets an action sequence and its state updates fit within a single LLM context, making joint multi-step transition a natural alternative to recursive prediction. Motivated by this, we develop a framework for learning future-predictive textual states and evolving them through structured multi-step updates over long horizons.
\section{Problem Setup}
\label{sec: prelim}

\subsection{World Modeling under Partial Observability}
\label{prelim_1: pomdp}

We model an interactive text environment as a POMDP \citep{kaelbling1998planning}. Let $s_t$ denote the latent environment state. After taking action $a_t$, the agent receives an outcome $x_{t+1}=(o_{t+1},r_{t+1})$, consisting of an observation and a step-level reward. Its available information at step $t$ is the interaction history $H_t=(x_1,a_1,\ldots,x_{t-1},a_{t-1},x_t)$.

In a POMDP, the belief state $p(s_t\mid H_t)$ is a sufficient statistic of the history \citep{kaelbling1998planning}. Predictive-state representations instead characterize state through action-conditional predictions of future observations \citep{littman2001predictive,singh2004predictive}. Following this view, we do not require the learned state $Z_t=f_{\mathrm{enc}}(H_t)$ to recover $s_t$ or its posterior; it need only preserve distinctions between histories that affect future observable outcomes.

For a prediction horizon $K$, an ideal predictive state satisfies
\begin{equation}
 p(x_{t+1:t+K}\mid H_t,a_{t:t+K-1})
 =
 p(x_{t+1:t+K}\mid Z_t,a_{t:t+K-1})
\end{equation}
for admissible future action sequences. This allows histories with different surface forms to share a state when they induce the same future observable dynamics. In text environments, such a state must integrate distributed events, track action-induced changes, and discard information that is no longer predictively relevant. In our experiments, the decoded prediction target is the observation component $o$ of $x$; we predict rewards only where an environment reports them as part of $o$ (e.g., Jericho's game score), rather than modeling $r$ as a separate output.

Given $\hat Z_t=f_{\mathrm{enc}}(H_t)$, subsequent prediction is conditioned on the history only through $\hat Z_t$. For any horizon $k\leq K$,
\begin{equation}
H_t
\xrightarrow{\mathrm{encoding}}
\hat Z_t
\xrightarrow[a_{t:t+k-1}]{\mathrm{transition}}
\hat Z_{t+k}
\xrightarrow{\mathrm{decoding}}
\hat o_{t+k}.
\end{equation}
Thus, information needed for future prediction must be represented in $\hat Z_t$ and preserved as the state evolves. The remaining design question is how to parameterize this latent transition; as detailed next, recursive and structured-reasoning transitions additionally expose the intermediate states $\hat Z_{t+1:t+k-1}$ along the way, in different forms.

\subsection{Transition Framework Variants}
\label{prelim_3: transition types}

We consider three transition variants over a $k$-step action sequence: \emph{recursive} prediction, \emph{direct} non-recursive prediction, and our \emph{structured reasoning} (SR) formulation. The \emph{recursive} model uses $k$ successive one-step calls, whereas the latter two process the initial state and full action sequence in a single call; they differ in whether intermediate structure is exposed.

The recursive formulation initializes $\hat Z^{\mathrm{rec}}_t=\hat Z_t$ and repeatedly applies a shared one-step transition:
\begin{equation}
\hat Z^{\mathrm{rec}}_{t+j}
=
f^{\mathrm{rec}}_{\mathrm{trans}}
\left(\hat Z^{\mathrm{rec}}_{t+j-1},a_{t+j-1}\right),
\qquad j=1,\ldots,k.
\end{equation}
Each predicted state is therefore fed back into the next transition call.

The two non-recursive variants instead condition jointly on $\hat Z_t$ and the complete action sequence. A \emph{direct} transition predicts only the horizon-$k$ endpoint:
\begin{equation}
\hat Z^{\mathrm{dir}}_{t+k}
=
f^{\mathrm{dir}}_{\mathrm{trans}}
\left(\hat Z_t,a_{t:t+k-1}\right).
\end{equation}
Our structured-reasoning formulation instead emits a sequence of sparse state edits within the same call:
\begin{equation}
\hat\Delta_{1:k}
=
f^{\mathrm{SR}}_{\mathrm{trans}}
\left(\hat Z_t,a_{t:t+k-1}\right),
\end{equation}
and the endpoint $\hat Z^{\mathrm{SR}}_{t+k}$ is obtained mechanically by applying these edits to $\hat Z_t$, rather than generated directly. These edits are never re-entered as inputs to a new transition call, so they can be supervised without reintroducing recursive feedback. Thus, recursive prediction composes one-step predictions, direct prediction jumps to the endpoint, and SR remains non-recursive while exposing the intermediate trajectory for supervision. Section~\ref{sec: method} details its training.

\subsection{Empirical Probes of Compounding Error}
\label{prelim_4: empirical probes}

Recursive and non-recursive transitions differ in how errors can propagate across the horizon. Under recursive rollout, an erroneous intermediate state becomes input to later calls; under a non-recursive transition, no predicted state is re-entered in this way. We probe this difference from two complementary angles.

First, we compare matched recursive and non-recursive models while holding backbone, data, and training recipe fixed within each stage. The non-recursive transition achieves higher long-horizon accuracy across stages, and the gap persists as training improves (Section~\ref{sec:ablation}). Second, we inject a fabricated fact into the latent state and track its effect with horizon. Recursive rollout continues to carry the perturbation, whereas its effect attenuates under structured reasoning (Section~\ref{sec:state-probes}). Together, these probes motivate a non-recursive transition with explicit intermediate structure.

\section{Methodology}
\label{sec: method}

Section~\ref{prelim_1: pomdp} defines the encode--transition--decode factorization, and Section~\ref{prelim_3: transition types} distinguishes recursive, direct, and structured-reasoning (SR) transitions. We instantiate encoding, transition, and decoding with a shared autoregressive policy $\pi_\theta$ using stage-specific prompts. Our method has two components: (i) a replayable sparse-edit representation that exposes the internal multi-step trajectory without recursive feedback, and (ii) trajectory-level training that supervises both intermediate and terminal predictive quality.

\subsection{Structured Multi-Step State Transition}
\label{method: transition}

Given an encoded state $\hat Z_t$ and future actions $a_{t:t+k-1}$, the SR transition produces only action-aligned edits $\hat\Delta_{1:k}$; no state, intermediate or final, is generated directly. Each $\hat\Delta_i$ specifies the typed changes induced by action $a_{t+i-1}$; unaffected facts need not be regenerated. Appendix~\ref{app:latent-format} details the concrete textual format of $\hat Z_t$ and $\hat\Delta_i$, with examples in Appendix~\ref{app:examples}.

Every state along the trajectory, including the horizon-$k$ endpoint, is obtained by deterministic replay of these edits:
\begin{equation}
\tilde Z_{t+i}
=
\mathrm{Apply}\!\left(\hat Z_t,\hat\Delta_{1:i}\right),
\qquad i=1,\ldots,k.
\label{eq:replay}
\end{equation}
Here, $\hat Z_t$ denotes the encoded state the model conditions on, whereas $\tilde Z_{t+i}$ is mechanically reconstructed by a deterministic programmatic \textsc{Apply} operator; the language model never generates a state directly beyond $\hat Z_t$ itself. Insertions, updates, and deletions are executed by code, so unchanged state is copied forward mechanically rather than regenerated, reducing an additional source of generation error. These reconstructed states are used only for supervision and are never fed back into the transition model.

\subsection{Training the Textual World Model}
\label{method: train}

The training data contain interaction histories, actions, and observed outcomes, but no reference textual states or edit trajectories. In particular, the latent transition is unobserved: much like intermediate reasoning traces, there is no ground-truth sequence of states or edits to imitate directly. We therefore train in two stages. First, we distill high-quality model-generated trajectories to initialize the structured latent computation. We then apply trajectory-level reinforcement learning, which optimizes those latent trajectories through their predictive consequences.

\subsubsection{Rejection-Sampled Distillation}
\label{method: sft}

A teacher policy $\pi_{\mathrm{tea}}$ generates complete encode--transition--decode traces on the training set. Because these latent traces are not ground truth, we retain only candidates whose terminal observation prediction exceeds a quality threshold and whose encoded state, edit sequence, and decoded observation satisfy the required formats. The accepted traces are then used as supervised distillation targets for the corresponding stages. Appendix~\ref{app:sft} gives the backbone, distillation data construction, and training objective.

This distillation stage provides a structured initialization rather than a final learning objective: it teaches the policy to produce coherent latent states and edit trajectories before reinforcement learning evaluates and improves them based on predictive outcomes. Without such an initialization, malformed or uninformative latent trajectories can dominate early rollouts and obscure the learning signal.

\subsubsection{Trajectory-Level Reinforcement Learning}
\label{method: rl}

\textbf{Rollout-based credit assignment.}
Because the textual states and edits are discrete sampled sequences, we optimize them through their predictive consequences using Group Relative Policy Optimization (GRPO)~\citep{shao2024deepseekmath}. We sample a rollout tree with branching factors $(b_{\mathrm{enc}}, b_{\mathrm{trans}}, b_{\mathrm{dec}})$. Each complete rollout receives a scalar reward, and each interior node is assigned the mean reward of its descendants. Siblings under the same parent form the group for computing group-relative advantages, yielding stage-specific learning signals for encoding, transition, and decoding from the same trajectory-level outcome. Appendix~\ref{app:rl} gives the full reward definitions, including the variants used in the ablation of Section~\ref{sec:ablation}; Appendix~\ref{app:hyperparameters} lists training hyperparameters and compute.

\textbf{Predictive quality and predictive gain.}
For each decoded observation, $\mathrm{obs\_quality}$ measures agreement with the ground-truth observation using the benchmark-specific metric, while a format indicator $\mathrm{fmt}$ filters malformed outputs.

To encourage the learned state to support prediction rather than merely satisfy the output format, we compare the state-conditioned prediction with a matched history-conditioned pathway produced by the same policy, in the spirit of rewarding a model relative to a matched baseline rather than in absolute terms \citep{hatamizadeh2025rlp}. The latter predicts the observation directly from the full history and action and is used only to construct the reward. We define
\begin{equation}
\mathrm{obs\_quality\_gain}
=
\max\!\left(0,\;
\mathrm{obs\_quality\_policy}-\mathrm{obs\_quality\_base}
\right)
\Big/\, \tau_{\mathrm{gain}}.
\label{eq:predictive_gain}
\end{equation}
This term rewards states whose use improves predictive quality relative to the matched history-conditioned pathway, and contributes nothing when the state-conditioned prediction is no better than the baseline conditioned directly on raw history.

\textbf{Intermediate predictive supervision.}
The replayable edit sequence provides dense checkpoints along the transition. For each step $i$, we reconstruct $\tilde Z_{t+i}$ using \cref{eq:replay}, decode it into an observation, and evaluate it with the same predictive-quality metric as the terminal state. Averaging over intermediate steps gives $\mathrm{obs\_quality\_mid\_avg}$.

\textbf{Final reward.}
The trajectory reward is
\begin{equation}
\mathrm{reward} =
\mathrm{fmt}\cdot
\left(
\alpha_1\,\mathrm{obs\_quality}
+
\alpha_2\,\mathrm{obs\_quality\_gain}
+
\alpha_3\,\mathrm{obs\_quality\_mid\_avg}
\right).
\label{eq:final_reward}
\end{equation}
Here, $\alpha_1$, $\alpha_2$, and $\alpha_3$ weight terminal predictive fidelity, predictive gain, and intermediate predictive fidelity, respectively. Intermediate scores contribute to the reward of the jointly generated trajectory rather than defining separate optimization problems for individual edits.

\section{Experiments}
\subsection{Setup}
\label{sec:setup}

\textbf{Environments and Datasets}.
We study three partially observable interactive text environments: \textsc{ScienceWorld}~\citep{wang2022scienceworld}, \textsc{Jericho}~\citep{hausknecht19}, and \textsc{CEO-Bench}~\citep{chen2026ceo}. For each environment, we collect complete interaction trajectories using diverse policies and split them into disjoint training and test sets at the trajectory level before constructing prediction examples. Dataset construction and statistics are provided in Appendix~\ref{app:datasets}.

\textbf{Evaluation Protocol}.
To isolate rollout horizon from target-specific difficulty, we use fixed-target evaluation. For each target observation $o_T$, a horizon-$k$ prediction starts from the interaction history at step $T-k$ and conditions on the intervening actions $a_{T-k:T-1}$. Predictions across horizons therefore share the same target and differ only in rollout distance. On \textsc{ScienceWorld} and \textsc{CEO-Bench} we evaluate learned variants at $k=1,\ldots,10$; on \textsc{Jericho}, whose fixed-target set supports longer rollouts, at $k\in\{1,5,10,15,20\}$. Every table and ablation in this paper reports Avg.\ over a shared subsampled grid ($k\in\{1,3,5,7,9,10\}$ for \textsc{ScienceWorld}/\textsc{CEO-Bench}, all five horizons for \textsc{Jericho}) so that every model's average is comparable across tables. \textsc{ScienceWorld} is evaluated with Fact-F1 and BLEU (Appendices~\ref{app:fact-f1}--\ref{app:bleu}), and \textsc{Jericho} with BLEU and Score MAE, an absolute error on the game's internal score. For \textsc{CEO-Bench}, we use SE-score over cash, individual subscribers, enterprise seats, and open issues; Appendix~\ref{app:se-score} gives the full definition. All metrics are multiplied by $100$ for readability except Jericho Score MAE, which is left on its native scale.

\textbf{Baselines}.
We compare five learned world-model variants defined in Section~\ref{sec: prelim}: \texttt{obs-recursive}, \texttt{obs-direct}, \texttt{latent-recursive}, \texttt{latent-direct}, and \texttt{latent-SR}. The \texttt{obs-*} variants condition directly on the interaction history, whereas the \texttt{latent-*} variants operate on the learned predictive state. Comparing \texttt{obs-recursive} with \texttt{obs-direct} isolates recursive versus non-recursive prediction without a learned latent state; comparing \texttt{latent-direct} with the matched-reward \texttt{latent-SR} ablation isolates the structured transition, while the full \texttt{latent-SR} additionally evaluates the benefit of intermediate-state supervision (Section~\ref{sec:main}, Appendix~\ref{app:sft}); and \texttt{latent-recursive} provides the recursive latent counterpart. All learned variants use the same backbone, collected trajectories, data split, and evaluation protocol (Appendix~\ref{app:ho}). We additionally include GPT-5.2 (Appendix~\ref{app:gpt}) and, on \textsc{CEO-Bench}, Copy-last and learned time-series forecasters (Appendix~\ref{app:ts}).

For conservative comparison, we select the strongest GPT-5.2 prompt and time-series configuration directly on the test set. The selected time-series model is Chronos-2 with four target variables and 27 future-action covariates. It observes the same prefix and, where supported, future actions, but never future observations. These oracle-selected baselines are marked with $^\dagger$ in Table~\ref{tab:main-results} and Table~\ref{tab:ceo-results}; complete configurations are provided in Appendices~\ref{app:gpt} and~\ref{app:ts}.

\subsection{Main Results}
\label{sec:main}

\textbf{Overall performance}.
Across \textsc{ScienceWorld} (Table~\ref{tab:main-results}), \textsc{Jericho} (Table~\ref{tab:jericho-results}), and \textsc{CEO-Bench} (Table~\ref{tab:ceo-results}), \texttt{latent-SR} achieves the strongest average performance on every metric, with gains concentrated at longer horizons: at $h10$, it exceeds \texttt{obs-direct} by nearly 10 Fact-F1 and 13 BLEU points on \textsc{ScienceWorld}, and by about 7 SE-score points on \textsc{CEO-Bench}. On \textsc{Jericho}, \texttt{obs-recursive} leads at the shortest horizons ($h1$, $h5$) on both BLEU and Score MAE, but \texttt{latent-SR} overtakes it from $h10$ onward and attains the best average on both metrics.

\textbf{Qualitative case studies}.
Appendix~\ref{app:examples} gives paired examples illustrating these gains at the prediction level: \texttt{obs-direct} tends to reuse or extrapolate the most recent matching observation, while \texttt{latent-SR} reflects information that has since changed---retaining object availability after intervening failed actions in \textsc{Jericho}, updating the predicted room after an unobserved move in \textsc{ScienceWorld}, and tracking accumulated subscriber loss over sustained cancellations in \textsc{CEO-Bench} rather than returning a stale snapshot.

\textbf{Full-system comparison}.
The gap between \texttt{obs-direct} and \texttt{latent-SR} combines the latent state, structured transition, and training objectives, so Section~\ref{sec:ablation} separates these factors. \texttt{latent-direct} (same representation and training through predictive gain, but a direct transition) reaches 89.16/87.63 (Fact-F1/BLEU) on \textsc{ScienceWorld} and 83.53 SE-score on \textsc{CEO-Bench}---ahead of \texttt{obs-direct} but well below \texttt{latent-SR} (92.44/92.37 and 87.35). \texttt{latent-SR (Ours)} additionally uses the mid reward that \texttt{latent-direct} lacks; at the matched predictive-gain stage, \texttt{latent-SR}'s own ablations reach 86.58 SE-score on \textsc{CEO-Bench} (Table~\ref{tab:ceo-ablation}) and 91.91/91.10 (Fact-F1/BLEU) on \textsc{ScienceWorld} (Table~\ref{tab:sw-ablation}), so the transition architecture accounts for most of the gap over \texttt{latent-direct}, with the mid reward contributing a smaller remainder up to the full model's 87.35 and 92.44/92.37.

\textbf{Recursive prediction and other baselines}.
\texttt{obs-recursive} degrades more sharply with horizon than \texttt{obs-direct}; Section~\ref{sec:ablation} and the perturbation study in Section~\ref{sec:state-probes} examine whether this persists when representation and training are held fixed. GPT-5.2 degrades substantially at longer horizons despite test-set prompt selection; on \textsc{CEO-Bench}, Copy-last is competitive at short horizons but deteriorates as action-dependent changes accumulate, while the selected time-series model trails the learned world-model variants.

\begin{table*}[!htbp]
\centering
\caption{
CEO-Bench fixed-target long-horizon prediction.
}
\label{tab:ceo-results}
\resizebox{0.7\textwidth}{!}{%
\begin{tabular}{lccccccc}
\toprule
Method
& \(h1\) & \(h3\) & \(h5\)
& \(h7\) & \(h9\) & \(h10\) & Avg. \\
\midrule

GPT-5.2$^\dagger$
& \underline{92.67} & 86.70 & 81.31
& 78.11 & 72.80 & 69.46 & 80.17 \\

Copy-last
& 90.11 & 81.68 & 74.63
& 68.42 & 63.50 & 59.20 & 72.92 \\

TS model$^\dagger$
& 83.75 & 78.16 & 72.16
& 70.32 & 68.08 & 66.26 & 73.12 \\

obs-recursive
& 92.48 & 87.15 & \underline{84.34}
& 78.10 & 74.16 & 70.28 & 81.08 \\

obs-direct
& 89.04 & 85.86 & 82.72
& 80.48 & 78.19 & 74.81 & 81.85 \\

latent-recursive
& 92.24 & 87.73 & 82.81
& \underline{80.53} & 75.13 & 72.22 & 81.78 \\

latent-direct
& 92.25 & \underline{87.97} & 83.28
& 80.07 & \underline{79.13} & \underline{78.50} & \underline{83.53} \\

\textbf{latent-SR}
& \textbf{93.76} & \textbf{90.21} & \textbf{88.07}
& \textbf{85.99} & \textbf{84.33} & \textbf{81.73} & \textbf{87.35} \\

\bottomrule
\end{tabular}%
}
\end{table*}

\begin{table*}[!htbp]
\centering
\caption{
ScienceWorld fixed-target long-horizon prediction.
}
\label{tab:main-results}
\resizebox{\textwidth}{!}{%
\begin{tabular}{lcccccccccccccc}
\toprule
& \multicolumn{7}{c}{Fact-F1} & \multicolumn{7}{c}{BLEU} \\
\cmidrule(lr){2-8} \cmidrule(lr){9-15}
Method
& \(h1\) & \(h3\) & \(h5\) & \(h7\) & \(h9\) & \(h10\) & Avg.
& \(h1\) & \(h3\) & \(h5\) & \(h7\) & \(h9\) & \(h10\) & Avg. \\
\midrule

GPT-5.2$^\dagger$
& 82.13 & 83.55 & 79.22 & 76.34 & 68.67 & 68.21 & 76.35
& 59.67 & 60.27 & 59.29 & 57.26 & 52.71 & 52.67 & 56.98 \\

obs-recursive
& 94.45 & 90.53 & 84.70 & 79.42 & 76.76 & 73.66 & 83.25
& 89.16 & 86.01 & 79.91 & 72.44 & 64.98 & 63.25 & 75.96 \\

obs-direct
& \textbf{97.00} & \textbf{95.78} & \underline{91.19} & 87.14 & 82.91 & 79.62 & 88.94
& \underline{93.56} & \underline{93.15} & \underline{89.34} & 83.79 & 79.62 & 76.07 & 85.92 \\

latent-recursive
& 93.32 & 90.32 & 88.56 & 86.22 & 86.76 & 85.42 & 88.43
& 90.19 & 90.07 & 87.66 & 86.15 & 85.81 & 85.20 & 87.51 \\

latent-direct
& 91.45 & 90.74 & 88.19 & \underline{88.98} & \underline{87.78} & \underline{87.79} & \underline{89.16}
& 88.57 & 88.16 & 87.68 & \underline{87.30} & \underline{86.77} & \underline{87.28} & 87.63 \\

\textbf{latent-SR}
& \underline{96.36} & 93.01 & \textbf{92.53} & \textbf{93.09} & \textbf{90.27} & \textbf{89.38} & \textbf{92.44}
& \textbf{94.09} & \textbf{93.49} & \textbf{92.85} & \textbf{93.21} & \textbf{91.13} & 89.48 & \textbf{92.37} \\

\bottomrule
\end{tabular}%
}
\end{table*}

\begin{table*}[!htbp]
\centering
\caption{
Jericho fixed-target long-horizon prediction.
}
\label{tab:jericho-results}
\resizebox{\textwidth}{!}{%
\begin{tabular}{lcccccccccccc}
\toprule
& \multicolumn{6}{c}{BLEU} & \multicolumn{6}{c}{Score MAE (lower is better)} \\
\cmidrule(lr){2-7} \cmidrule(lr){8-13}
Method
& \(h1\) & \(h5\) & \(h10\) & \(h15\) & \(h20\) & Avg.
& \(h1\) & \(h5\) & \(h10\) & \(h15\) & \(h20\) & Avg. \\
\midrule

obs-recursive
& \textbf{38.0} & \textbf{25.2} & 20.3 & 18.6 & 15.1 & 23.4
& \textbf{0.217} & \textbf{1.039} & \underline{1.870} & 3.039 & 4.689 & 2.171 \\

obs-direct
& \underline{31.3} & 22.7 & 21.3 & 21.8 & 20.2 & 23.5
& \underline{0.246} & 1.453 & 2.311 & 2.696 & 3.072 & \underline{1.956} \\

latent-recursive
& 24.7 & 23.9 & \underline{23.6} & \textbf{23.9} & \underline{23.1} & \underline{23.9}
& 2.589 & 3.373 & 3.580 & 4.640 & 4.073 & 3.651 \\

latent-direct
& 22.4 & 22.6 & 21.7 & 21.8 & 22.0 & 22.1
& 1.927 & 2.357 & 2.567 & \underline{2.459} & \underline{2.737} & 2.409 \\

\textbf{latent-SR}
& 25.5 & \underline{25.1} & \textbf{24.2} & \underline{23.3} & \textbf{24.2} & \textbf{24.5}
& 1.276 & \underline{1.412} & \textbf{1.642} & \textbf{2.105} & \textbf{2.204} & \textbf{1.728} \\

\bottomrule
\end{tabular}%
}
\end{table*}

\subsection{Ablation Studies}
\label{sec:ablation}

Table~\ref{tab:ceo-ablation} isolates distillation and reward design for \texttt{latent-SR} on \textsc{CEO-Bench}; Table~\ref{tab:recursion-gap} (Appendix~\ref{app:recursion-gap}) separately compares \texttt{latent-SR} against the matched \texttt{latent-recursive} transition at each training stage. \textbf{Training.} Starting from distillation (83.49 average SE-score), trajectory RL raises it to 84.86 when initialized from distillation, whereas RL without this warm start drops to 61.12. Thus, teacher-distilled latent trajectories provide an important initialization for trajectory optimization in our setting.

\textbf{Reward design.} Adding predictive gain raises the average score to 86.58; adding intermediate-state rewards further improves it to 87.35. These gains are complementary to the transition architecture itself.

\begin{wraptable}{r}{0.7\textwidth}
\centering
\vspace{-10pt}
\caption{Ablation study on CEO-Bench.}
\label{tab:ceo-ablation}
\small
\resizebox{0.7\textwidth}{!}{%
\begin{tabular}{lccccccc}
\toprule
Variant
& \(h1\) & \(h3\) & \(h5\)
& \(h7\) & \(h9\) & \(h10\) & Avg. \\
\midrule

latent-SR (Distill)
& 92.59 & 88.18 & 85.21
& 81.19 & 78.58 & 75.21 & 83.49 \\

latent-SR (GRPO, no Distill)
& 91.96 & 85.68 & 68.36
& 49.83 & 41.89 & 28.98 & 61.12 \\

latent-SR (Distill+GRPO)
& 92.82 & \underline{89.58} & 87.53
& 84.00 & 78.49 & 76.76 & 84.86 \\

\quad \(+\) predictive gain
& \underline{93.01} & 89.37 & \underline{87.96}
& \underline{85.05} & \underline{83.14} & \underline{80.98}
& \underline{86.58} \\

\quad \(+\) mid reward
& \textbf{93.76} & \textbf{90.21} & \textbf{88.07}
& \textbf{85.99} & \textbf{84.33} & \textbf{81.73} & \textbf{87.35} \\

\bottomrule
\end{tabular}%
}
\end{wraptable}

\textbf{Recursion gap across training stages.}\label{sec:recursion-gap} We additionally compare matched recursive and non-recursive variants, holding backbone, data, and training recipe fixed at three stages: SFT, Distill, and Distill+GRPO (no predictive gain or intermediate-state reward at any stage). The average SE-score gap is $+0.8$ for \texttt{obs-direct} versus \texttt{obs-recursive}, $+2.3$ for distilled \texttt{latent-SR} versus \texttt{latent-recursive}, and widens to $+3.1$ at Distill+GRPO (Table~\ref{tab:recursion-gap}, Appendix~\ref{app:recursion-gap}), suggesting recursive feedback is a source of degradation that trajectory RL does not close. \texttt{latent-recursive} at this matched stage reaches 81.78 average SE-score, the row reported in Table~\ref{tab:ceo-results}.

\textbf{Delta reasoning.} We also compare sparse delta reasoning against full-state reasoning, which explicitly regenerates the complete intermediate state after each action. On a shared 500-example \textsc{ScienceWorld} subset, delta reasoning improves Fact-F1 from 0.499 to 0.556 with Qwen3-4B and 0.597 to 0.625 with GPT-5.2, supporting sparse state updates over repeatedly regenerating unchanged state. Free-form thinking (unstructured chain-of-thought text instead of a state-schema-constrained format) scores lower on the same subset with Qwen3-4B (0.465) and with GPT-5.2 (0.606).

\subsection{Controlled State Perturbation}
\label{sec:state-probes}

To probe error propagation directly, we intervene on the encoded state in \textsc{CEO-Bench} while holding the future action sequence fixed. Given encoder output $z_t$, we inject the same fabricated fact across architectures and report
$\Delta(k)=\mathrm{SE}_{\mathrm{clean}}(k)-\mathrm{SE}_{\mathrm{pert}}(k)$
under the fixed-target protocol of Section~\ref{sec:setup}; larger values indicate greater damage. The \textsc{relevant} perturbation inserts a fabricated $-\$5{,}000$/week exogenous cash shock contradicting the realized trajectory; the \textsc{irrelevant} condition inserts a latency distractor outside the four scored fields. Evaluation covers the same held-out companies as the main \textsc{CEO-Bench} results (Appendix~\ref{app:dataset-stats}), at all ten horizons under greedy decoding.

\textbf{Perturbation damage persists under recursion but is smaller under latent-SR.}
The relevant perturbation harms every model, whereas the irrelevant distractor has near-zero effect (Table~\ref{tab:perturb}), showing the intervention is selective rather than generic corruption. For \texttt{latent-recursive}, it causes an 8--14 point SE-score loss across horizons (11.1 average); the average loss is 5.4 for \texttt{latent-SR} after distillation and 1.7 for the full model.

\texttt{latent-recursive} repeatedly turns predicted intermediate states into inputs to later transition calls, so an erroneous fact can persist across calls; \texttt{latent-SR} instead conditions a single multi-step generation on the original $z_t$ without recursively re-entering reconstructed states. This does not by itself identify the internal mechanism, but is consistent with reduced propagation through recursive feedback.

\textbf{Trajectory RL further improves robustness.}
Holding the \texttt{latent-SR} architecture fixed, trajectory training reduces mean perturbation damage from 5.4 after distillation to 1.7 in the full model, while clean SE-score increases from 83.2 to 87.1 on the same subset. Thus the robustness gain is not obtained by simply degrading the model's clean prediction quality.

\begin{table*}[t]
\centering
\caption{
State perturbation on CEO-Bench. Higher $\Delta(k)$ = greater damage.
}
\label{tab:perturb}
\resizebox{0.9\textwidth}{!}{%
\begin{tabular}{llccccccccccc}
\toprule
Metric & Model & $h1$ & $h2$ & $h3$ & $h4$ & $h5$ & $h6$ & $h7$ & $h8$ & $h9$ & $h10$ & Avg. \\
\midrule
\multirow{3}{*}{\shortstack[l]{$\Delta(k)$\\\textsc{relevant}}}
& latent-recursive & 8.0 & 9.9 & 13.3 & 12.9 & 11.3 & 8.5 & 11.9 & 10.0 & 13.6 & 12.0 & 11.1 \\
& latent-SR (Distill) & 6.1 & 5.6 & 5.9 & 6.7 & 5.9 & 5.3 & 5.8 & 4.1 & 5.3 & 3.0 & 5.4 \\
& latent-SR & \textbf{3.0} & \textbf{2.5} & \textbf{1.1} & \textbf{2.5} & \textbf{2.7} & \textbf{1.9} & \textbf{0.9} & \textbf{2.0} & \textbf{0.8} & \textbf{0.1} & \textbf{1.7} \\
\midrule
\multirow{3}{*}{\shortstack[l]{$\Delta(k)$\\\textsc{irrelevant}}}
& latent-recursive & $-$1.2 & 1.1 & 1.4 & 0.2 & $-$0.4 & $-$0.7 & $-$1.0 & $-$0.2 & $-$0.3 & 0.4 & $-$0.1 \\
& latent-SR (Distill) & 0.0 & 0.5 & 0.6 & 0.0 & $-$0.6 & 1.1 & 1.1 & $-$0.1 & $-$0.5 & 0.8 & 0.3 \\
& latent-SR & $-$0.1 & $-$0.1 & $-$0.5 & 0.0 & 0.4 & 0.3 & $-$1.2 & 0.3 & $-$0.4 & 0.2 & $-$0.1 \\
\midrule
\multirow{3}{*}{\shortstack[l]{Clean\\SE-score}}
& latent-recursive & 90.2 & 88.3 & 86.4 & 84.7 & 81.4 & 77.2 & 77.1 & 73.0 & 71.9 & 69.5 & 80.0 \\
& latent-SR (Distill) & 92.0 & 88.7 & 87.3 & 87.3 & 84.1 & 81.3 & 82.1 & 78.4 & 76.0 & 74.5 & 83.2 \\
& latent-SR & \textbf{93.1} & \textbf{90.7} & \textbf{88.8} & \textbf{89.6} & \textbf{88.2} & \textbf{85.6} & \textbf{84.8} & \textbf{86.1} & \textbf{83.2} & \textbf{80.7} & \textbf{87.1} \\
\bottomrule
\end{tabular}%
}
\end{table*}

\subsection{Counterfactual Action Sensitivity}
\label{sec:counterfactual-winrate}

Aggregate accuracy does not show whether a model's predictions actually track the supplied action sequence. We test this on \textsc{CEO-Bench} by forking a shared history into three branches (\emph{aggressive growth}, \emph{cost cutting}, \emph{resample control}) that diverge by horizon 10, and asking each model to predict one branch's outcome given only that branch's actions.

\begin{wraptable}{r}{0.62\textwidth}
\centering
\vspace{-10pt}
\caption{Company-clustered counterfactual action sensitivity (two-sided tests; Appendix~\ref{app:counterfactual}).}
\label{tab:adjm}
\small
\begin{tabular}{lccc}
\toprule
Method & Adj.M (cl.) & Sign-test $p$ & Perm.\ $p$ \\
\midrule
obs-recursive & $-0.040$ & $<0.001$ & --- \\
obs-direct & $-0.036$ & $<0.001$ & --- \\
Chronos-2 (FT) & $+0.000$ & $0.690$ & $0.209$ \\
\textbf{latent-SR} & $\mathbf{+0.024}$ & $\mathbf{<0.001}$ & $\mathbf{<0.001}$ \\
\bottomrule
\end{tabular}
\end{wraptable}

We report an \emph{adjusted margin}: how much closer a prediction is to its own branch's outcome than to the alternatives, minus the same margin from a branch-blind null (Chronos-2, zero-shot), isolating discrimination due to the branch's actions rather than ground-truth geometry. We test significance two ways: a sign test and a permutation test, both applied at the company level since predictions from the same company are not independent. Full statistical detail is in Appendix~\ref{app:counterfactual}.

\texttt{latent-SR} is the only method significant under \emph{both} tests ($p<0.001$ each), consistent with Section~\ref{sec:state-probes}. Chronos-2 (FT) shows no discrimination on either test. \texttt{obs-recursive} and \texttt{obs-direct} are significantly negative under the sign test; we omit their permutation $p$-value here because that statistic is unreliable for these two rows (Appendix~\ref{app:counterfactual} explains why). Appendix~\ref{app:counterfactual} shows this robustness is specific to the full reward design (every earlier \texttt{latent-SR} stage is null under both tests), gives a worked case study (Table~\ref{tab:cf-case-study}), and reports a second, independent test that reaches the same conclusion without referencing Chronos-2 at all.

\section{Conclusion}
\label{sec:conclusion}

We propose latent-SR, a non-recursive textual world-model transition that predicts sparse state edits in a single call and reconstructs intermediate states deterministically for supervision. Across \textsc{ScienceWorld}, \textsc{Jericho}, and \textsc{CEO-Bench}, it achieves the strongest average long-horizon prediction performance, with larger gains at longer horizons. Controlled comparisons further show reduced sensitivity to propagated state errors and improved action-conditioned counterfactual prediction. These results support structured, trajectory-explicit latent transitions as a promising alternative to recursive one-step rollout.
Our experiments focus on textual, partially observable simulated environments; whether the same factorization benefits visual or continuous world models, or improves downstream closed-loop planning, remains open.

\clearpage

\subsection*{AI use statement}

We used generative AI tools (large language model assistants) to help draft and edit portions of the manuscript text and to assist with writing and debugging experiment and analysis code. All AI-assisted text and code were reviewed and edited by the authors, and all reported experimental results were independently verified against the underlying logs and evaluation outputs before inclusion. Separately from writing assistance, GPT-5.2 is also used as an experimental artifact in this work: it is one of the evaluated baseline models (Section~\ref{sec:setup}, Appendix~\ref{app:gpt}) and serves as the teacher policy for rejection-sampled distillation (Section~\ref{method: sft}); this usage is a research contribution under study, not a writing aid, and is documented as such throughout the paper. We take responsibility for the final content of this work, including text, claims, and artifacts produced with the aid of generative AI.

\subsection*{Ethics statement}

This work studies textual world models in simulated interactive environments (\textsc{ScienceWorld}, \textsc{Jericho}, and \textsc{CEO-Bench}) and does not involve human subjects, personally identifiable information, or crowdsourced data collection. All environments and models used are either publicly available research benchmarks or synthetic simulations; no proprietary or sensitive data is used. \textsc{CEO-Bench} is a synthetic business-simulation environment and does not involve real companies, individuals, or financial data. We are not aware of any direct harms, dual-use concerns, or fairness/bias issues specific to this work beyond those generally associated with training and evaluating large language models.

\subsection*{Reproducibility statement}

We describe our environments, dataset construction, and evaluation protocol in Section~\ref{sec:setup} and Appendix~\ref{app:datasets}, including trajectory collection policies, train/test splits, and the fixed-target evaluation procedure. The model architecture and training procedure (distillation and trajectory-level reinforcement learning, including the full reward formulation) are described in Section~\ref{sec: method}, with backbone models, hyperparameters, and compute details in Appendix~\ref{app:hyperparameters}. Baseline implementations, including the baselines conditioned directly on raw history, GPT-5.2 prompting protocol, and numeric forecasting baselines, are detailed in Appendix~\ref{app:baselines}, and the exact prompt templates used for every architecture are given in Appendix~\ref{app:prompts}. Evaluation metric definitions (Fact-F1, BLEU, SE-score) are given in Appendix~\ref{app:metrics}, and the adjusted-margin counterfactual evaluation methodology is detailed in Appendix~\ref{app:counterfactual}.



\clearpage

\bibliography{references}
\bibliographystyle{plainnat}

\clearpage

\appendix


\section{Dataset Construction}
\label{app:datasets}

\subsection{Environments}
\label{app:environments}

\textbf{\textsc{ScienceWorld}}~\citep{wang2022scienceworld}. 30 task families spanning physics, chemistry, and biology (e.g. \texttt{boil}, \texttt{melt}, \texttt{freeze}, \texttt{chemistry-mix}, \texttt{find-animal}, \texttt{grow-plant}, \texttt{measure-melting-point}, \texttt{mendelian-genetics}, \texttt{power-component}), each instantiated with several object/room variations. Observations are free-text room and object descriptions (e.g. room contents, container/device state, substance state and measurements). The action space is open natural-language commands (e.g. ``go to hallway'', ``pour red paint into wood cup''), not a fixed discrete set. Episodes are capped at 40 steps.

\textbf{\textsc{Jericho}}. Seven text-adventure games: \texttt{zork1}, \texttt{enchanter}, \texttt{sorcerer}, \texttt{ludicorp}, \texttt{deephome}, \texttt{dragon}, \texttt{balances} (\texttt{zork3} is excluded from the fixed-target evaluation set because its rollouts do not reach a late enough base step to support the longest horizons). Observations are Z-machine text output, including a bracketed running score tag (e.g. \texttt{[SCORE: 10/51]}). The action space is open natural-language interactive-fiction commands (e.g. ``take coin'', ``north'', ``examine grimoire''), not a fixed discrete set.

\textbf{\textsc{CEO-Bench}}.
Each step corresponds to one simulated week, in which the agent files actions for a simulated company. Each observation reports the company's business state, including cash, individual subscribers, enterprise seats, and open issues.
Beyond these four scored ``level'' fields, each observation additionally reports 12 ``dynamics'' fields describing the week's flows and service health (new individual/enterprise leads, new individual subscribers, new enterprise seats, cancellations, upgrades, downgrades, p95 latency, error rate, outage and overload indicators), plus 4 ``config'' fields that carry forward the agent's own action choices from the previous step (price, capacity, ops spend, dev spend) --- these config fields are predicted and fed back into history but are not scored. The action space is the CEO agent's choice of these same 4 levers each week (continuous values, parsed from free text), i.e. 4 action types per step.

\subsection{Trajectory Collection}
\label{app:collection}

For each environment, we collect complete interaction trajectories using diverse policies.
\textbf{\textsc{ScienceWorld}}: two base policies, \texttt{oracle} (an optimal task solver) and \texttt{methodical} (a systematic but non-optimal explorer), each also run in a noisy variant that mixes in a fraction of random actions (\texttt{epsilon\_greedy\_0.1}: 10\% random deviation from the oracle; \texttt{noisy\_*}: added action noise on top of either base policy), for 6 realized policy variants total. The training split has 540 trajectories (oracle 62, noisy\_oracle 123, methodical 67, noisy\_methodical 121, epsilon\_greedy\_0.1 63, noisy\_epsilon\_greedy\_0.1 104) and the test split has 60 (same 6 policies), with a maximum episode length of 40 steps.
\textbf{\textsc{Jericho}}: five policies --- \texttt{expert} (walkthrough replay), \texttt{random\_walk} (uniform random over valid actions), \texttt{epsilon\_greedy} (walkthrough with a 30\% chance of a random action at each step), \texttt{explore\_then\_oracle} (5 random warm-up steps, then walkthrough), and \texttt{failure\_modes} (walkthrough with early truncation to also cover failure states). The evaluation set uses 7 games $\times$ 4 independent seeded rollouts each (28 trajectories), with a maximum episode length of 60 steps.
\textbf{\textsc{CEO-Bench}}: trajectories are full simulated company histories (one action-observation pair per week) under the CEO agent's own policy, run for up to roughly 52 weeks; 165 companies are used for training and 67 held out for test.

\subsection{Data Split and Example Construction}
\label{app:split}

Trajectories are split into disjoint training and test sets at the trajectory level before any prediction example is constructed, so histories from the same trajectory never appear in both splits. Each prediction example is indexed by a trajectory \(i\), a base step \(t\), and a horizon \(k\). Its input consists of the interaction history up to step \(t\) and the action sequence \((a_t,\ldots,a_{t+k-1})\), and its target is the observation \(o_{t+k}\).
For fixed-target evaluation, test examples are grouped by the target step \(T=t+k\), so that predictions at all horizons for the same target are evaluated against the same \(o_T\).
A target step \(T\) is kept for the fixed-target test set only if every horizon \(k=1,\ldots,H\) (\(H=10\) for ScienceWorld/CEO-Bench, \(H=20\) for Jericho) has a valid base step \(t=T-k\) within the same trajectory, so the test set forms a rectangular anchor-by-horizon grid with no missing cells; this is why the number of test targets (Appendix~\ref{app:dataset-stats}) is smaller than the number of eligible base steps. Training examples are not required to satisfy this full-horizon-coverage constraint, since only the test protocol needs every horizon scored against the same target; horizons \(k\) for training examples are instead sampled non-uniformly per environment (CEO-Bench's training pool, for instance, is weighted toward \(k\in\{1,5,10\}\)).

\subsection{Dataset Statistics}
\label{app:dataset-stats}

\begin{table}[h]
\centering
\small
\caption{Dataset statistics.}
\label{tab:dataset-stats}
\resizebox{\columnwidth}{!}{%
\begin{tabular}{lcccccc}
\toprule
Environment & \#Train traj. & \#Test traj. & \#Train ex. & \#Test targets & Avg. traj. len. & Avg. obs. tokens \\
\midrule
\textsc{ScienceWorld} & 540 & 60 & 17{,}573 & 298 & 32.5 & 40 \\
\textsc{Jericho}      & --- & 28\rlap{\textsuperscript{a}} & 4{,}131 & 415 & 28.6 & 33 \\
\textsc{CEO-Bench}    & 165 & 67\rlap{\textsuperscript{b}} & 4{,}740 & 128 & 52 & 14{,}500\rlap{\textsuperscript{c}} \\
\bottomrule
\end{tabular}%
}
\end{table}

\footnotesize
\textsuperscript{a}~7 games $\times$ 4 independent seeded rollouts each; \((\text{game\_id}, \text{seed})\) is the trajectory identity. Train-side trajectory count is not separately tracked (train examples are built directly from rolled-out transitions, deduplicated at the window level; see Appendix~\ref{app:collection}).
\textsuperscript{b}~67 held-out companies (simulated-week trajectories). The 128 reported test targets are fixed-target anchors $T$ for which every horizon $k=1,\ldots,10$ has a valid base step within the same trajectory (Appendix~\ref{app:split}); each anchor is scored at all 10 horizons, giving $128\times10=1{,}280$ (anchor, horizon) prediction rows in total, drawn from the 67 held-out trajectories (multiple anchors per company). The state-perturbation experiment of Section~\ref{sec:state-probes} evaluates these same 67 held-out companies, selecting one fixed-target anchor per company and scoring it at all ten horizons (rather than the multiple anchors per company used above); it uses the identical held-out set, not a separate sample, and shares no companies with the counterfactual set of Appendix~\ref{app:counterfactual}, which is drawn from a different simulator run.
\textsuperscript{c}~Median token count of the rendered TSV-format history block fed to the encoder (max 29{,}463, well under the 32{,}768 training context; transition-stage inputs are considerably shorter, median $\approx$2{,}300 tokens).

\section{Evaluation Metrics}
\label{app:metrics}

\subsection{Fact-F1}
\label{app:fact-f1}

Facts are extracted from observation text with an environment-specific parser (\texttt{delphi.metrics.eval\_observation}) into predicate strings (e.g. room identity, container contents, device/substance state, numeric measurements). Predicted and reference fact sets are matched greedily in three passes: (1) exact predicate-and-argument matches; (2) numeric predicates matched under a small relative tolerance (5\%, with a small absolute floor, so near-identical measurements are not penalized for formatting/rounding differences); (3) remaining reference facts with placeholder arguments matched structurally against remaining unmatched predictions. The matched count is the true positives; precision is true positives over the number of predicted facts, recall is true positives over the number of reference facts, and Fact-F1 is their harmonic mean, averaged over all test examples at each horizon.

\subsection{BLEU}
\label{app:bleu}

We compute BLEU between predicted and target observations on \textsc{ScienceWorld} and \textsc{Jericho}.
We use NLTK's \texttt{sentence\_bleu} (sentence-level, not corpus-level), computed per example and then averaged. Predicted and reference text are lowercased and whitespace-tokenized after stripping chat-template special tokens and markdown code fences. BLEU-4 is used by default, but the effective n-gram order is capped at \(\min(4, |\hat{o}|, |o|)\) so that short one-line observations (common in Jericho, e.g. ``Taken.'') are not forced to a score of zero purely because they are shorter than 4 tokens; NLTK's \texttt{SmoothingFunction} method 1 is applied.

\subsection{SE-score}
\label{app:se-score}

\textsc{CEO-Bench} observations describe the state of a simulated company, and the quantities that matter most for evaluation are numeric. Token-overlap metrics such as BLEU compare numbers as strings and do not reflect how far a predicted value is from the true one. We therefore evaluate \textsc{CEO-Bench} predictions with SE-score, which measures the relative error on four key state fields.

Let \(y=(y_1,\ldots,y_4)\) denote the true values of cash, individual subscribers, enterprise seats, and open issues at the target step, and let \(\hat{y}=(\hat{y}_1,\ldots,\hat{y}_4)\) denote the corresponding values extracted from the predicted observation. SE-score is defined as
\begin{equation}
\mathrm{SE}(\hat{y}, y) = 1 - \frac{1}{4}\sum_{i=1}^{4} \min\!\left(1, \frac{|\hat{y}_i - y_i|}{\max(100, |y_i|)}\right).
\label{eq:se-score}
\end{equation}
Each term is the relative error of one field. The floor of 100 in the denominator keeps fields with small true values, such as the number of open issues, from producing disproportionately large relative errors: when \(|y_i| < 100\), the error is measured relative to 100. The cap at 1 bounds each field's contribution, so a single badly mispredicted field cannot dominate the score. SE-score therefore lies in \([0,1]\) and equals 1 only when all four fields are predicted exactly.
If a prediction cannot be parsed at all, every field's error term is set to 1 (i.e. \(\mathrm{SE}=0\) for that example) rather than the example being dropped from the mean, so that unparseable generations at long horizons cannot quietly inflate the average by shrinking the denominator. If a prediction parses but an individual field cannot be recovered as a number, that field's error term alone is set to 1, while the remaining fields are still scored normally.
We report the average SE-score over all test examples at each horizon, multiplied by 100.

\section{Implementation Details}
\label{app:implementation}

\subsection{Latent State Representation}
\label{app:latent-format}

On ScienceWorld and CEO-Bench, the state $\hat Z$ is a single flat JSON object of free-text key-value pairs: keys are short descriptive strings chosen by the model (no fixed schema), and values combine facts restated from the observation history with inferred latent variables (e.g. a trend, a driver, an unstated consequence) that are useful for future prediction but not directly stated. Edits $\hat\Delta_i$ are sparse flat JSON objects containing only the keys whose value changed as a result of the corresponding action; omitted keys are unchanged (\cref{eq:replay}'s \textsc{Apply} operator merges the edit into the previous state). On Jericho, the state is instead a nested JSON object (e.g. \texttt{\{"agent": \{"location": "kitchen"\}, "objects": \{...\}, "score": 5\}}), with no fixed schema beyond validity as JSON; an edit is a list of structured operations \texttt{\{"op": "set"|"remove", "path": "<dot.separated.path>", "value": ...\}} addressing into the state by dot-separated key path. See Appendix~\ref{app:prompts} for the exact prompts that produce and consume this representation, and Appendix~\ref{app:examples} for worked examples.

\subsection{Supervised Fine-Tuning}
\label{app:sft}

All learned variants (\texttt{obs-recursive}, \texttt{obs-direct}, \texttt{latent-recursive}, \texttt{latent-direct}, \texttt{latent-SR}) use Qwen3~\citep{yang2025qwen3} as the backbone.
All learned variants, including the latent-SR SFT/distillation stage, are fine-tuned from Qwen3-4B-Instruct-2507. Distillation data for latent-SR is constructed by rejection sampling: a teacher policy (GPT-5.2, best-of-$N$ sampled) generates complete encode--transition--decode traces on the training set (Section~\ref{method: sft}), producing a per-action edit list at the transition stage (Appendix~\ref{app:prompts}), and candidates are filtered to those whose terminal observation prediction and intermediate stage formats meet a quality threshold. The retained traces' free-text reasoning is stripped, leaving each stage's final structured output (state / per-action edit list / observation text) as the supervised target. Training uses the standard next-token cross-entropy SFT objective over these targets, rendered as multi-turn chat-format messages matching the encode/transition/decode call structure.

\textbf{\texttt{latent-direct}.} \texttt{latent-direct} shares the same encoder, decoder, textual state representation, and backbone as \texttt{latent-SR}, but replaces the structured transition with a single-call predictor of the endpoint state $Z_{t+k}$ directly, without exposing intermediate edits. Its distillation data is teacher-generated in the same way, except the transition-stage target is only the final state (no intermediate edit sequence, since none is produced). \texttt{latent-direct} is trained through the predictive-gain stage using the same terminal objectives as the corresponding \texttt{latent-SR} variant; because it never exposes intermediate states, it does not receive the intermediate-state (mid) reward. Comparing \texttt{latent-direct} against the matched-reward (Distill+GRPO+predictive gain) \texttt{latent-SR} ablation therefore isolates the structured transition itself, while the reported full \texttt{latent-SR} additionally evaluates the benefit of intermediate-state supervision.

\subsection{Reinforcement Learning}
\label{app:rl}

Our full model is further optimized with trajectory-level rewards consisting of a predictive-gain reward and an intermediate-state reward. We use GRPO~\citep{shao2024deepseekmath} (Section~\ref{method: rl}); \cref{eq:final_reward} gives the reward at the level of the paper's notation ($\mathrm{obs\_quality}$, $\mathrm{obs\_quality\_gain}$, $\mathrm{obs\_quality\_mid\_avg}$, weighted by $\alpha_1,\alpha_2,\alpha_3$). Concretely, $\mathrm{obs\_quality}$ (\texttt{obs\_score} in code) is the same BLEU/Fact-F1 blend used for evaluation (a weighted combination, weight 0.5 by default, of BLEU and Fact-F1 against the ground-truth observation, or SE-score on CEO-Bench); $\mathrm{obs\_quality\_gain}$ (the predictive-gain term, \texttt{ig}) is $\mathrm{obs\_quality\_policy}-\mathrm{obs\_quality\_base}$, where the baseline prediction is produced by the same policy conditioned directly on history and action (a frozen, pre-computed baseline, scored once rather than resampled live at each RL step) rather than the learned state, and is computed only at the terminal state. The intermediate-state reward, $\mathrm{obs\_quality\_mid\_avg}$, reuses the same $\mathrm{obs\_quality}$ computation at each replayed intermediate checkpoint $\tilde Z_{t+i}$ (\cref{eq:replay}) and averages the result, so intermediate states are supervised by the same terminal metric rather than a separate objective.

\textbf{Reward variants for ablation}.
The rows of Table~\ref{tab:ceo-ablation} correspond to the following settings of \cref{eq:final_reward}: \emph{Distill+GRPO} sets $\alpha_2=\alpha_3=0$, so the reward reduces to terminal predictive fidelity only; \emph{+ predictive gain} turns on $\alpha_2$ and \cref{eq:predictive_gain}, $\mathrm{obs\_quality\_gain}=\max(0, \mathrm{ig})/\tau_{\mathrm{gain}}$, which contributes nothing when the state-conditioned prediction is no better than the baseline conditioned directly on raw history rather than actively penalizing it; \emph{+ mid reward} additionally turns on $\alpha_3$ (the intermediate-state term $\mathrm{obs\_quality\_mid\_avg}$) on top of the predictive-gain variant, giving our full model.

\subsection{Hyperparameters and Compute}
\label{app:hyperparameters}

All stages use Qwen3-4B-Instruct-2507 as the backbone (Appendix~\ref{app:sft}), trained with the \texttt{verl} framework. Table~\ref{tab:hparams} reports the realized hyperparameters for each stage; the GRPO row for each environment applies uniformly across every reward/architecture variant trained at that stage, not a per-variant setting: this includes Table~\ref{tab:ceo-ablation}'s Distill+GRPO, $+$ predictive gain, and $+$ mid reward rows, every \texttt{latent-recursive} GRPO run in Table~\ref{tab:recursion-gap}, and \texttt{latent-direct} at its predictive-gain stage (Appendix~\ref{app:sft}). Decoding at evaluation is always greedy (temperature $=0$); training-time rollout diversity for GRPO instead comes from the branching factor (\(N\) independent action-conditioned continuations sampled per step, not from vLLM's \texttt{rollout.n}, which is fixed to 1).

\begin{table}[h]
\centering
\small
\caption{SFT and GRPO hyperparameters by environment and stage. Each GRPO row is shared by every reward/architecture variant trained at that stage in that environment. ``---'' marks a knob that does not apply to that stage (e.g. KL is disabled, not merely small).}
\label{tab:hparams}
\resizebox{\columnwidth}{!}{%
\begin{tabular}{llcccccl}
\toprule
Env. & Stage & LR & Batch & Epochs & GRPO $N$ & KL & Max length \\
\midrule
ScienceWorld & SFT / Distill & 1e-4 & 32--64 & 3 & --- & n/a & 8{,}192--32{,}768 \\
ScienceWorld & GRPO & 1e-5 & 16--32 & 3 & 4 & 0--0.01 & 2{,}048 resp.; prompt up to 32{,}768 \\
Jericho & SFT / Distill & 1e-5 & 32 & 3 & --- & n/a & 8{,}192 \\
Jericho & GRPO & 1e-6 & 16 & 3 & 4 & 0 & 8{,}192 / 2{,}048 \\
CEO-Bench & SFT / Distill & 1e-4 & 32 & 3 & --- & n/a & 32{,}768 \\
CEO-Bench & GRPO & 1e-5 & 16 & 3 & 4 & 0 & 32{,}768 / 2{,}048 \\
\bottomrule
\end{tabular}%
}
\end{table}

\textbf{Hardware.} SFT and the ScienceWorld/CEO-Bench GRPO stages run on 8$\times$A100-80GB; Jericho GRPO runs on 8$\times$H100-80GB. vLLM rollout GPU-memory utilization is tuned per run (0.35--0.6) to accommodate FSDP shard residency during simultaneous training and rollout. Per-step wall-clock time is approximately 3--4 minutes for ScienceWorld/CEO-Bench GRPO and 95--110 seconds for Jericho GRPO (a 300-step Jericho run takes roughly 8--9 hours end to end).

\section{Baseline Details}
\label{app:baselines}

\subsection{Raw-History Baselines}
\label{app:ho}

obs-recursive and obs-direct use the same backbone, collected trajectories, data split, and evaluation protocol as our model, and condition directly on the raw interaction history without the three-stage latent representation. obs-recursive is trained as a one-step predictor. At test time, it appends each predicted observation to the history and repeats the prediction \(k\) times. obs-direct is trained to predict \(o_{t+k}\) from the history and the full action sequence \((a_t,\ldots,a_{t+k-1})\) in a single call.
Their SFT hyperparameters do not differ from the settings in Appendix~\ref{app:hyperparameters} beyond the batch/epoch values already listed there per environment; no additional obs-recursive/obs-direct-specific tuning is applied.

\subsection{GPT-5.2}
\label{app:gpt}

We evaluate GPT-5.2~\citep{openai2025gpt5} with 3 prompt variants --- the same obs-recursive, obs-direct, and latent-SR prompts used for the corresponding learned architecture (Appendix~\ref{app:prompts}), substituting GPT-5.2 as the backend model with no further prompt engineering --- and report the variant with the highest test performance in Table~\ref{tab:main-results} (\textsc{ScienceWorld}) and Table~\ref{tab:ceo-results} (\textsc{CEO-Bench}). Results for all prompt variants are reported in Table~\ref{tab:ceo-all-baselines}; the prompts are listed in Appendix~\ref{app:prompts}.

\subsection{Numeric Forecasting Baselines}
\label{app:ts}

\textbf{Copy-last} ignores actions and predicts that the state at the target step equals the most recent observed state.

\textbf{Learned time-series models}.
We evaluate Chronos-2 and TimesFM~3.0 under different target-variable sets and covariate settings: Chronos-2 with 4 targets plus 27 future-action covariates, Chronos-2 with 4 targets and no covariates, Chronos-2 with 20 targets and no covariates, and TimesFM~3.0 with 20 targets (multivariate, no covariate support). All models use the same observed prefix and, where supported, the known future actions as covariates, but never future observations. We report the configuration with the highest mean test SE-score across horizons in Table~\ref{tab:ceo-results}. This selects Chronos-2 with four target variables and 27 future-action covariates.
Results for all configurations are reported in Table~\ref{tab:ceo-bench-4b}.

\begin{table}[t]
\centering
\caption{CEO-Bench results across architectures and training stages --- SE-score. gain: predictive-gain reward (\cref{eq:predictive_gain}); mid: additional intermediate-state reward.}
\label{tab:ceo-bench-4b}
\resizebox{\columnwidth}{!}{%
\begin{tabular}{llccccccc}
\toprule
Architecture & Training & h1 & h3 & h5 & h7 & h9 & h10 & avg \\
\midrule
\multicolumn{9}{l}{\textit{TS baselines (numeric only)}} \\
\midrule
persistence & Copy-last-value & 0.9011 & 0.8168 & 0.7463 & 0.6842 & 0.6350 & 0.5920 & 0.7292 \\
Chronos-2 & 4v + 27 action cov & 0.8375 & 0.7816 & 0.7216 & 0.7032 & 0.6808 & 0.6626 & 0.7312 \\
TimesFM 3.0 & 20v multivariate & 0.8869 & 0.8160 & 0.7319 & 0.6564 & 0.6170 & 0.5756 & 0.7140 \\
Chronos-2 & 4v, no covariates & 0.8302 & 0.7796 & 0.7135 & 0.6865 & 0.6536 & 0.6363 & 0.7166 \\
Chronos-2 & 20v, no covariates & 0.8184 & 0.7748 & 0.7118 & 0.6929 & 0.6568 & 0.6371 & 0.7153 \\
\midrule
obs-recursive & GPT-5.2 & 0.9258 & 0.8648 & 0.8100 & 0.7799 & 0.7182 & 0.6689 & 0.7946$^\dagger$ \\
obs-direct & GPT-5.2 & 0.9266 & 0.8620 & 0.8060 & 0.7668 & 0.7135 & 0.6795 & 0.7924$^\dagger$ \\
latent-SR & GPT-5.2 & 0.9267 & 0.8670 & 0.8131 & 0.7811 & 0.7280 & 0.6946 & 0.8018$^\dagger$ \\
\midrule
obs-recursive & SFT & 0.9248 & 0.8715 & 0.8434 & 0.7810 & 0.7416 & 0.7028 & 0.8108 \\
obs-direct & SFT & 0.8904 & 0.8586 & 0.8272 & 0.8048 & 0.7819 & 0.7481 & 0.8185 \\
latent-recursive & Distill & 0.9233 & 0.8710 & 0.8174 & 0.7819 & 0.7450 & 0.7326 & 0.8119 \\
latent-SR & Distill & 0.9259 & 0.8818 & 0.8521 & 0.8119 & 0.7858 & 0.7521 & 0.8349 \\
latent-SR & GRPO, no Distill & 0.9196 & 0.8568 & 0.6836 & 0.4983 & 0.4189 & 0.2898 & 0.6112 \\
latent-recursive & Distill+GRPO & 0.9224 & 0.8773 & 0.8281 & 0.8053 & 0.7513 & 0.7222 & 0.8178 \\
latent-SR & Distill+GRPO & 0.9282 & 0.8958 & 0.8753 & 0.8400 & 0.7849 & 0.7676 & 0.8486 \\
latent-direct & Distill+GRPO+gain & 0.9225 & 0.8797 & 0.8328 & 0.8007 & 0.7913 & 0.7850 & 0.8353 \\
latent-SR & Distill+GRPO+gain & 0.9301 & 0.8937 & 0.8796 & 0.8505 & 0.8314 & 0.8098 & 0.8659 \\
latent-SR & Distill+GRPO+gain+mid & \textbf{0.9376} & \textbf{0.9021} & \textbf{0.8807} & \textbf{0.8599} & \textbf{0.8433} & \textbf{0.8173} & \textbf{0.8735} \\
\bottomrule
\end{tabular}%
}
\end{table}

\section{Additional Results}
\label{app:additional-results}

\subsection{ScienceWorld Ablation Study}
\label{app:sw-full}

Table~\ref{tab:sw-ablation} reports the ScienceWorld counterpart of the CEO-Bench ablation in Table~\ref{tab:ceo-ablation} (Section~\ref{sec:ablation}): full per-horizon Fact-F1 and BLEU as trajectory-level RL and predictive-gain shaping are incrementally added on top of self-distilled \texttt{latent-SR} training.

\begin{table*}[t]
\centering
\caption{
\textbf{ScienceWorld ablation study measured by Fact-F1 and BLEU.}
All variants use the latent-SR transition unless otherwise specified.
Each row incrementally adds trajectory-level reinforcement learning and predictive-gain shaping to self-distilled Distill training.
Best and second-best results among the ablation variants are shown in bold and underlined, respectively (independently per metric).
}
\label{tab:sw-ablation}
\resizebox{\textwidth}{!}{%
\begin{tabular}{lcccccccccccccc}
\toprule
& \multicolumn{7}{c}{Fact-F1} & \multicolumn{7}{c}{BLEU} \\
\cmidrule(lr){2-8} \cmidrule(lr){9-15}
Variant
& \(h1\) & \(h3\) & \(h5\) & \(h7\) & \(h9\) & \(h10\) & Avg.
& \(h1\) & \(h3\) & \(h5\) & \(h7\) & \(h9\) & \(h10\) & Avg. \\
\midrule

latent-SR (Distill)
& 91.39 & 90.99 & 88.82 & 90.04 & 87.23 & 85.99 & 89.08
& 84.22 & 86.15 & 81.51 & 82.13 & 79.37 & 77.23 & 81.77 \\

latent-SR (Distill+GRPO)
& 91.87 & \underline{93.31} & 91.30 & 89.39 & \underline{90.08} & 87.23 & 90.53
& 91.17 & 91.46 & 91.18 & 89.44 & 88.53 & 86.98 & 89.79 \\

\quad \(+\) predictive gain
& \underline{95.79} & \textbf{93.60} & \textbf{92.84} & \underline{91.36} & 89.98 & \underline{87.91} & \underline{91.91}
& \underline{93.46} & \underline{92.29} & \textbf{93.00} & \underline{91.49} & \underline{89.21} & \underline{87.16} & \underline{91.10} \\

\textbf{latent-SR (Ours)}
& \textbf{96.36} & 93.01 & \underline{92.53} & \textbf{93.09} & \textbf{90.27} & \textbf{89.38} & \textbf{92.44}
& \textbf{94.09} & \textbf{93.49} & \underline{92.85} & \textbf{93.21} & \textbf{91.13} & \textbf{89.48} & \textbf{92.37} \\

\bottomrule
\end{tabular}%
}
\end{table*}

\subsection{Recursive versus Non-Recursive Transition Across Training Stages}
\label{app:recursion-gap}

Table~\ref{tab:recursion-gap} reports the full per-horizon results underlying the comparison in Section~\ref{sec:recursion-gap}: on \textsc{CEO-Bench}, we hold backbone, data, and training recipe fixed at each of three stages---raw history (obs, trained with ordinary supervised fine-tuning, SFT), self-distillation (Distill), and self-distillation followed by trajectory-level RL (Distill+GRPO)---and vary only the transition parameterization between recursive and non-recursive (obs-direct at the raw-history stage; latent-SR at the Distill and Distill+GRPO stages).

\begin{table*}[t]
\centering
\caption{
\textbf{The recursion gap across training stages, on \textsc{CEO-Bench}.}
Each pair holds backbone, data, and training recipe fixed and varies only the transition:
recursive versus non-recursive. Raw-history pairs compare obs-recursive to obs-direct; latent pairs compare latent-recursive to latent-SR. SE-score, \(\times 100\).
Gap = non-recursive \(-\) recursive, averaged over horizons.
}
\label{tab:recursion-gap}
\resizebox{\textwidth}{!}{%
\begin{tabular}{llcccccccc}
\toprule
Training stage & Transition
& \(h1\) & \(h3\) & \(h5\)
& \(h7\) & \(h9\) & \(h10\) & Avg. & Gap \\
\midrule

\multirow{2}{*}{Raw history (obs)}
& non-recursive (direct) & 89.04 & 85.86 & 82.72 & 80.48 & 78.19 & 74.81 & 81.85 & \multirow{2}{*}{\(+0.8\)} \\
& recursive & 92.48 & 87.15 & 84.34 & 78.10 & 74.16 & 70.28 & 81.08 & \\
\midrule

\multirow{2}{*}{Distill (SFT)}
& non-recursive (SR) & 92.59 & 88.18 & 85.21 & 81.19 & 78.58 & 75.21 & 83.49 & \multirow{2}{*}{\(+2.3\)} \\
& recursive & 92.33 & 87.10 & 81.74 & 78.19 & 74.50 & 73.26 & 81.19 & \\
\midrule

\multirow{2}{*}{Distill + trajectory RL}
& non-recursive (SR) & 92.82 & 89.58 & 87.53 & 84.00 & 78.49 & 76.76 & \textbf{84.86} & \multirow{2}{*}{\(+3.1\)} \\
& recursive & 92.24 & 87.73 & 82.81 & 80.53 & 75.13 & 72.22 & 81.78 & \\

\bottomrule
\end{tabular}%
}
\end{table*}

\subsection{Counterfactual Branch Prediction on \textsc{CEO-Bench}}
\label{app:counterfactual}

\textbf{Branch construction}.
From a shared interaction history and encoder state at step \(t_0\), we generate three 10-week continuations under different policy directives: \emph{aggressive growth} (spend to acquire), \emph{cost cutting} (preserve cash), and \emph{resample control} (no additional steering). Each directive is used only to generate the branch's actions and simulated outcomes; the directive itself is never shown to the model. This counterfactual set comprises 60 source trajectories, generated from a separate simulator run and disjoint from the main \textsc{CEO-Bench} train/test split (Appendix~\ref{app:dataset-stats}). We refer to it as the early-branching set, since the three branches are constructed so that their realized outcomes have diverged by horizon 10 for every source company; an earlier, non-early-branching version of this set had a much smaller fraction of sources diverge by horizon 10 and is not used for the results reported here.

\textbf{Prediction task}.
At each horizon \(k \in \{1, 3, 7, 10\}\) weeks past \(t_0\), the model is given the shared encoder state at \(t_0\) and \emph{one} branch's filed actions for weeks \(t_0+1,\ldots,t_0+k\), and predicts that branch's state at \(t_0+k\). It never sees the other two branches' actions or outcomes. Predicted and true states are compared with SE-score (\cref{eq:se-score}).

\textbf{Raw win rate and its confound}.
For a prediction conditioned on branch \(b\), a three-way win is recorded when the prediction is at least as close to its own outcome as to both alternatives:
\begin{equation}
\operatorname{win}(\hat{y}_b) = \mathbb{I}\!\left[\mathrm{SE}(\hat{y}_b, y_b) \geq \max_{b'\neq b} \mathrm{SE}(\hat{y}_b, y_{b'})\right].
\end{equation}
Without ties, the nominal chance rate is \(\frac{1}{3}\); exact ties are counted in the model's favor. This raw win rate is confounded by ground-truth geometry: when two branches' realized outcomes happen to be close to each other, any prediction landing near that shared region wins against both regardless of whether it used that branch's actions. Because this inflates win rate for every model roughly equally, raw win rate for all evaluated models in our data falls within a narrow \(57\)--\(65\%\) band, including for a branch-blind baseline that predicts identically for every branch.

\textbf{Adjusted margin}.
To isolate discrimination attributable to the branch's actions, we define, for prediction \(\hat y\) made for branch \(b\),
\begin{equation}
\mathrm{SE}_{\mathrm{own}} = \mathrm{SE}(\hat y, y_b), \qquad
\mathrm{SE}_{\mathrm{alt}} = \operatorname*{mean}_{b'\neq b}\, \mathrm{SE}(\hat y, y_{b'}), \qquad
\mathrm{margin} = \mathrm{SE}_{\mathrm{own}} - \mathrm{SE}_{\mathrm{alt}},
\end{equation}
so that a positive margin indicates a prediction closer to its own branch. We then subtract the margin obtained by a branch-blind null baseline (Chronos-2, zero-shot, which predicts identically across branches) on the same \((\text{source}, \text{horizon}, \text{branch})\) triples:
\begin{equation}
\mathrm{Adj.M} = \mathrm{margin} - \mathrm{margin}_{\mathrm{null}}.
\end{equation}
This removes the free margin any prediction receives purely from branch ground-truth geometry, isolating the margin attributable to actually conditioning on the branch's actions. We report the median \(\mathrm{Adj.M}\) over all \((\text{source}, \text{horizon}, \text{branch})\) triples with \(h \geq 5\), together with a two-sided sign test on the fraction of positive margins; \(\mathrm{Adj.M}=0\) by construction for the null baseline itself.

\textbf{Company-clustered test.} The \((\text{source}, \text{horizon}, \text{branch})\) triples used above share only the 60 underlying source companies of the counterfactual set, so treating each triple as an independent observation in the sign test risks understating the true variance. We therefore additionally aggregate all triples for a given source company to a single median \(\mathrm{Adj.M}\), yielding one clustered statistic per company (\(n=60\)), and run a two-sided sign test at this level. We further validate significance with a company-level \emph{sign-flip permutation test} (not a bootstrap, since no data are resampled with replacement): for each resample, we draw one random sign (\(\pm 1\), probability \(\frac{1}{2}\)) per company and apply it to every triple belonging to that company, preserving the within-company correlation structure (10,000 resamples), and the two-sided permutation \(p\)-value is the fraction of resamples whose \(|\text{median clustered statistic}|\) meets or exceeds the observed \(|\text{median clustered statistic}|\). Columns ``Adj.M (cl.)'' and sign-test/permutation \(p\) in Table~\ref{tab:adjm-full} report this company-level statistic.

The median-of-medians statistic underlying this permutation test is well-behaved when most companies carry a nonzero clustered value, but degenerates when a majority of the per-company medians tie at exactly the same value (most commonly zero): the observed \(|\text{median}|\) then sits at that tied value, resamples that also keep a majority at the same tied value trivially satisfy the meets-or-exceeds criterion, and the permutation \(p\)-value is driven toward 1 regardless of how large the untied companies' margins are. A mean-based statistic does not share this failure mode, since it is not dominated by a tied majority the way a median is (the within-model action-permutation test described later in this section uses a mean-based statistic for exactly this reason). \texttt{obs-recursive} and \texttt{obs-direct}'s permutation \(p=1.000\) alongside a highly significant negative sign test is consistent with this degeneracy; we report their sign-test result, which does not share this failure mode, and omit an unverified permutation \(p\)-value for those two rows in the main text (Table~\ref{tab:adjm}).

Table~\ref{tab:adjm-full} reports adjusted margin across the full training and reward ablation, extending the summary in Table~\ref{tab:adjm}, including Chronos-2 fine-tuned (FT) on \textsc{CEO-Bench} with 27 future-action covariates (27act). Discrimination is null (median \(\approx 0\), non-significant) for every \texttt{latent-SR} variant prior to intermediate-state supervision, indicating that the significant positive margin of the full model is specifically attributable to that reward component rather than to distillation, trajectory RL, or predictive gain alone. At the company-clustered level, \texttt{latent-SR} remains significant under both the sign test and the permutation test, with a comfortably nonzero median clustered statistic. \texttt{obs-recursive} and \texttt{obs-direct} remain significantly negative under the clustered sign test; their permutation \(p=1.000\) is not interpreted as evidence against action sensitivity, per the degeneracy noted above.

\begin{table}[t]
\centering
\caption{
\textbf{Counterfactual action sensitivity, full ablation.}
Extends Table~\ref{tab:adjm} with every \texttt{latent-SR} training stage. Company-clustered statistics (``cl.'') are primary; per-triple statistics are shown for reference.
}
\label{tab:adjm-full}
\small
\begin{tabular}{lcccccc}
\toprule
Method & Adj.M (cl.) & Sign-test $p$ (cl.) & Perm. $p$ (cl.) & Median Adj.M & Sign-test $p$ \\
\midrule
obs-recursive & $-0.040$ & $<0.001$ & $1.000$ & $-0.112$ & $<0.001$ \\
obs-direct & $-0.036$ & $<0.001$ & $1.000$ & $-0.096$ & $0.003$ \\
Chronos-2 (27act, FT) & $+0.000$ & $0.690$ & $0.209$ & $-0.000$ & $0.940$ \\
\midrule
latent-SR (Distill) & $-0.000$ & $0.711$ & $0.770$ & $-0.000$ & $0.404$ \\
latent-SR (Distill+GRPO) & $+0.000$ & $0.845$ & $0.983$ & $-0.000$ & $0.287$ \\
latent-SR (+ predictive gain) & $+0.000$ & $0.265$ & $0.135$ & $+0.000$ & $0.404$ \\
\textbf{latent-SR (Ours)} & $\mathbf{+0.024}$ & $\mathbf{<0.001}$ & $\mathbf{<0.001}$ & $+0.048$ & $0.012$ \\
\bottomrule
\end{tabular}
\end{table}

\begin{table*}[t]
\centering
\caption{
\textbf{Complete CEO-Bench baseline results.}
We report fixed-target SE-score multiplied by 100; Avg.\ is computed over these six horizons for every row, matching Table~\ref{tab:ceo-results}.
The learned time-series and GPT-5.2 configurations reported in the main table are selected directly on the test set.
A star marks configurations retained in the main table.
}
\label{tab:ceo-all-baselines}
\resizebox{0.7\textwidth}{!}{%
\begin{tabular}{lccccccc}
\toprule
Candidate
& \(h1\) & \(h3\) & \(h5\) & \(h7\) & \(h9\) & \(h10\) & Avg. \\
\midrule

\multicolumn{8}{l}{\textit{Numeric forecasting baselines}} \\
\midrule
Copy-last$^\star$
& 90.11 & 81.68 & 74.63 & 68.42 & 63.50 & 59.20 & 72.92 \\

Chronos-2 (4 targets + 27 action covariates)$^\star$
& 83.75 & 78.16 & 72.16 & 70.32 & 68.08 & 66.26 & 73.12 \\

TimesFM 3.0 (20 targets)
& 88.69 & 81.60 & 73.19 & 65.64 & 61.70 & 57.56 & 71.40 \\

Chronos-2 (4 targets, no covariates)
& 83.02 & 77.96 & 71.35 & 68.65 & 65.36 & 63.63 & 71.66 \\

Chronos-2 (20 targets, no covariates)
& 81.84 & 77.48 & 71.18 & 69.29 & 65.68 & 63.71 & 71.53 \\

\midrule
\multicolumn{8}{l}{\textit{GPT-5.2 prompt candidates}} \\
\midrule
obs-recursive prompt
& 92.58 & 86.48 & 81.00 & 77.99 & 71.82 & 66.89 & 79.46 \\

obs-direct prompt
& 92.66 & 86.20 & 80.60 & 76.68 & 71.35 & 67.95 & 79.24 \\

latent-SR prompt$^\star$
& 92.67 & 86.70 & 81.31 & 78.11 & 72.80 & 69.46 & 80.18 \\

\bottomrule
\end{tabular}%
}
\end{table*}

\textbf{Case study.} Table~\ref{tab:cf-case-study} shows one concrete branching example (source company, week 2, \$811K cash, 139 customers) illustrating what the adjusted margin measures. All three branches share the same actions for the first week, so every method's $h{=}1$ cash forecast is nearly identical; the branches then diverge sharply by $h{=}10$ under their respective directives, most visibly under \emph{aggressive growth} (cash falls to \$250K as spending outpaces new revenue). \texttt{latent-SR} tracks this divergence in the correct direction and stays within \$1.5K--\$1.7K of ground truth on the two flatter branches, and within \$345K on the sharply diverging \emph{aggressive growth} branch, whereas \texttt{Chronos-2} (branch-blind) does not condition on the branch-specific future actions and therefore produces the same \$815K--\$850K trajectory across all three branches; this drifts increasingly far from all three ground truths as they diverge, missing the true \emph{aggressive growth} outcome by \$599K at $h{=}10$.

\textbf{Within-model action-permutation test.} The Adj.M analysis above isolates action-conditioned discrimination by subtracting a branch-blind null baseline (Chronos-2). As a complementary check that does not rely on any external model as a reference point, we also test each checkpoint against itself: for prediction $\hat y$ made conditioned on branch $b$'s own actions, we compute $\mathrm{own}(b) = \mathrm{SE}(\hat y, y_b)$, and separately generate a prediction conditioned on a different branch $b'$'s actions from the same starting state, $\mathrm{swapped}(b,b') = \mathrm{SE}(\hat y_{b'}, y_b)$, scored against $b$'s own ground truth. The statistic $\delta = \mathrm{own}(b) - \mathrm{swapped}(b,b')$ asks whether the model's own prediction changes appropriately when only the conditioning actions change, holding the starting state fixed. We aggregate $\delta$ to a per-company mean (rather than median, since a majority of companies have an exact-zero $\delta$ for a sizable fraction of checkpoints and horizons --- the model produces a scoring-identical prediction regardless of which branch's actions it is given, a manifestation of the same under-responsiveness to specific divergent actions noted elsewhere in this appendix, and a median statistic would degenerate under this tied majority exactly as noted above) and run both a two-sided sign test and a company-level sign-flip permutation test (10,000 resamples), analogous to the Adj.M company-clustered test above.

\texttt{latent-SR} reaches company mean $\delta=+0.0056$ with sign test $p=6.96\times10^{-5}$ and permutation $p=0.0012$, while the SFT baseline and an earlier Distill+GRPO stage without the intermediate-state reward remain non-significant ($p \geq 0.40$ on both tests). This reproduces, with a test that never references Chronos-2 or any other model, the same qualitative pattern as the Adj.M ablation: significant action sensitivity is specific to the full reward design, not present at earlier training stages. A majority of companies have an exact-zero $\delta$ for a sizable fraction of checkpoints and horizons, cutting the effective sample size that carries information.

\begin{table}[t]
\centering
\caption{
\textbf{Counterfactual case study.} One source company (\$811K cash, 139 customers at the branch point), forked into three directives. Chronos-2 does not condition on the branch-specific future actions and therefore produces the same forecast across all three branches.
}
\label{tab:cf-case-study}
\small
\begin{tabular}{lccc}
\toprule
Branch & $h{=}1$ & $h{=}5$ & $h{=}10$ \\
\midrule
\multicolumn{4}{@{}l}{\emph{Continue as before}} \\
\quad Ground truth & \$809,362 & \$808,968 & \$806,608 \\
\quad \texttt{latent-SR} & \$810,839 & \$807,989 & \$805,080 \\
\quad \texttt{Chronos-2} & \$815,059 & \$818,769 & \$849,645 \\
\midrule
\multicolumn{4}{@{}l}{\emph{Aggressive growth}} \\
\quad Ground truth & \$761,984 & \$626,447 & \$250,289 \\
\quad \texttt{latent-SR} & \$810,839 & \$733,300 & \$595,300 \\
\quad \texttt{Chronos-2} & \$815,059 & \$818,769 & \$849,645 \\
\midrule
\multicolumn{4}{@{}l}{\emph{Cost cutting}} \\
\quad Ground truth & \$810,110 & \$809,779 & \$807,803 \\
\quad \texttt{latent-SR} & \$810,839 & \$808,468 & \$806,103 \\
\quad \texttt{Chronos-2} & \$815,059 & \$818,769 & \$849,645 \\
\bottomrule
\end{tabular}
\end{table}

\section{Prompt Templates}
\label{app:prompts}

GPT-5.2 is evaluated with the same prompt templates used to train/query the corresponding learned architecture (below), substituting GPT-5.2 as the backend model; no GPT-5.2-specific prompt engineering is applied beyond this substitution. Below, each architecture's I/O contract is given once in canonical form -- consistent with how the main text defines the five variants (Section~\ref{sec: method}) -- followed by what differs per environment. \texttt{obs-recursive} and \texttt{obs-direct} condition directly on the raw interaction history; \texttt{latent-recursive}, \texttt{latent-direct}, and \texttt{latent-SR} all share the same ENCODE and DECODE stages and differ only in how TRANSITION is called.

\subsection{obs-recursive}
\label{app:prompt-obs-recursive}

One call per step; the predicted observation is spliced back into the history before the next call, repeated \(k\) times to reach horizon \(k\).
\begin{verbatim}
Input:
history = (o_0, a_0, o_1, a_1, ..., o_t)
a_t

Output:
o_{t+1}
\end{verbatim}
The system prompt frames the task as ``you are a world model for [the environment]; given observation history and an action, predict the next observation,'' instructed to match the environment's observation style and output only the observation text; the user prompt renders the interleaved history followed by the action to condition on. ScienceWorld and Jericho use the same template, differing only in the environment name and style reference in the system prompt.

\subsection{obs-direct}
\label{app:prompt-obs-direct}

One call per fixed-target example: the model is shown the entire future action sequence and must predict only the terminal observation, without producing or being shown intermediate observations.
\begin{verbatim}
Input:
history = (o_0, a_0, o_1, a_1, ..., o_t)
[a_t, a_{t+1}, ..., a_{t+k-1}]

Output:
o_{t+k}
\end{verbatim}
The user prompt renders the interleaved history followed by the full future action sequence in order, and instructs the model to reason over but not emit the intermediate steps, outputting only the observation after the last listed action. ScienceWorld and Jericho again share this contract and differ only in environment-specific phrasing.

\subsection{latent-recursive, latent-direct, and latent-SR}
\label{app:prompt-latent-sr}

All three latent variants share the same ENCODE and DECODE stages:
\begin{verbatim}
ENCODE
Input:  history = (o_0, a_0, o_1, a_1, ..., o_t)
Output: Z_t

DECODE
Input:  Z_{t+1}, a_t
Output: o_{t+1}
\end{verbatim}
$Z_t$ is the environment's latent state (Appendix~\ref{app:latent-format}); ENCODE maps the full history to $Z_t$, and DECODE maps a state together with the action that produced it to the corresponding observation text. They differ only in how TRANSITION maps $Z_t$ and the future action sequence to $Z_{t+k}$:
\begin{verbatim}
latent-recursive (one call per action, chained k times to reach horizon k):
Input:  Z_i, a_i          Output: Z_{i+1}

latent-direct (one call, no exposed intermediate states):
Input:  Z_t, [a_t, ..., a_{t+k-1}]     Output: Z_{t+k}

latent-SR (one call, per-action edits exposed):
Input:  Z_t, [a_t, ..., a_{t+k-1}]
Output: {
  "steps": [{"state_delta": Delta_1}, ..., {"state_delta": Delta_k}]
}
(Z_{t+k} is never generated; it is obtained by mechanically applying
Delta_1, ..., Delta_k to Z_t.)
\end{verbatim}
\texttt{latent-direct} isolates the representation effect (latent state vs. raw history) at a fixed direct-call mechanism, matched against \texttt{obs-direct}; \texttt{latent-recursive} isolates the same representation effect under recursive feedback, matched against \texttt{obs-recursive}; \texttt{latent-SR} instead exposes the per-action edit sequence $\hat\Delta_{1:k}$ (\cref{eq:replay}), with the endpoint $\tilde Z_{t+k}$ mechanically replayed rather than generated. The reported \texttt{latent-SR} additionally uses the intermediate-state (mid) reward that \texttt{latent-direct} does not (Section~\ref{method: rl}), so the \texttt{latent-direct} vs.\ \texttt{latent-SR} comparison in Table~\ref{tab:ceo-results} reflects both the transition mechanism and this reward term. Section~\ref{sec:main} isolates the transition mechanism's own contribution using Table~\ref{tab:ceo-ablation}'s matched reward stage (Distill+GRPO+predictive gain).

Environments differ only in what $Z_t$ and each $\Delta_i$ look like, and in the environment-specific parts of ENCODE/DECODE:

\textbf{ScienceWorld}: $Z_t$ is a single flat JSON object of free-text key-value pairs (facts plus inferred latent variables, no fixed schema); each $\Delta_i$ is a sparse flat JSON object containing only the keys that changed as a direct result of action $a_{t+i-1}$. DECODE matches ScienceWorld's observation style.

\textbf{Jericho}: $Z_t$ is a JSON object (nested key-values, e.g. \texttt{\{"agent": \{"location": "kitchen"\}, "objects": \{...\}, "score": 5\}}); each $\Delta_i$ is a list of structured edit operations, \texttt{\{"op": "set"|"remove", "path": "<dot.separated.path>", "value": ...\}}, addressing into $Z_t$ by dot-separated key path. For \texttt{latent-SR}, the model emits one structured edit entry per action across the whole action sequence (Section~\ref{method: transition}); the endpoint state is then obtained mechanically by replaying these edits, never generated directly. The teacher traces used to build distillation data include free-form reasoning before this structured output, but that reasoning is stripped before training (Appendix~\ref{app:sft}), so the trained model does not reproduce it at inference. DECODE is a domain-agnostic renderer that infers each game's exact observation phrasing from the in-context history (Jericho spans 50+ heterogeneous games with no shared observation grammar) and appends the running score as a \texttt{[SCORE: n/max]} tag, since Jericho's game score is otherwise not recoverable from decoded text alone.

\textbf{CEO-Bench}: $Z_t$ is a single flat JSON object of latent variables $z_t$ (one entry per field or cross-field driver, each a short free-text description of its rate of change, trend, driver, and forward implication) merged with the observed fields $o_t$; each $\Delta_i$ is a sparse flat JSON object of the (observed-field or latent-variable) keys that changed in week $t+i$. There is no separate DECODE call: the mechanically replayed endpoint state already contains the observable fields needed to score SE-score directly, so ENCODE and TRANSITION are the only two calls, sent as single user-role messages with no system turn, matching the format of its SFT training rows verbatim.

\section{Case Study Examples}
\label{app:examples}

We examine paired examples to illustrate when a latent-based world model improves over an obs-direct predictor (Tables~\ref{tab:jericho_cases}--\ref{tab:ceo_cases}). The shared pattern is that the target requires maintaining or updating a state, rather than directly reusing a past observation. In Jericho, the latent model retains an oven's availability despite intervening failed actions, preserves access to a grimoire in an opened container, and predicts the destination of a navigation sequence where obs-direct returns the preceding location. In ScienceWorld, it tracks movement through an unobserved action window, replaces an outdated kitchen description with the correct hallway observation, and predicts a successful drawer-opening action where obs-direct returns an invalid-action message. In CEO, the examples span subscriber depletion, sustained decline, and net growth: the latent predictions better reflect the accumulated change in subscriber stock, whereas obs-direct either retains outdated values or predicts an unsupported collapse.

These cases illustrate a conditional advantage, rather than establishing that longer histories or more state-changing actions inherently favor latent models. In the accompanying paired analysis, the mean latent-minus-obs-direct difference is $+0.084$ BLEU for Jericho examples with low target--history lexical overlap ($R<0.3$), $+0.092$ Fact-F1 for ScienceWorld windows containing a room transition, and $+0.059$ in CEO's four-field score when no historical snapshot attains a copy score of $0.9$. Here, lexical $R$ is the maximum token-set F1 between the target and any historical observation; CEO copy scores use the same four-field metric as model evaluation. These are retrospective, dataset-specific strata, not a common scale or evidence that retrievability is the strongest factor in every dataset. The nine examples are selected illustrations, not independent significance tests or causal evidence about the model's internal computation. Remaining errors also limit the interpretation: Jericho's correct room text can coexist with an incorrect game score, and CEO's improved terminal stocks do not imply accurate intermediate flows or all-field forecasts.

\begin{table*}[t]
\centering
\small
\setlength{\tabcolsep}{5pt}
\renewcommand{\arraystretch}{1.12}
\caption{Three paired Jericho examples. obs-direct is SFT; latent is \texttt{latent-SR}. BLEU is the saved terminal observation BLEU-4, excluding the game-score tag. Text excerpts below are taken from the original outputs.}
\label{tab:jericho_cases}
\begin{tabularx}{\textwidth}{@{}p{0.13\textwidth}>{\raggedright\arraybackslash}X>{\raggedright\arraybackslash}X@{}}
\toprule
 & \textbf{Raw-history} & \textbf{Latent-based} \\
\midrule
\multicolumn{3}{@{}p{\textwidth}@{}}{\textbf{J1: Retaining an earlier room state.} \texttt{enchanter\_step33\_k10\_s201}; $t=33$, $k=10$.} \\
\multicolumn{3}{@{}p{\textwidth}@{}}{\textbf{Context.} Observation $o_{14}$ describes an oven in the shack; $o_{17}$ again places the agent inside it. Later observations include failed movements. The forecast window includes \texttt{take all}, \texttt{D}, \texttt{E}, \texttt{E}, \texttt{put down book}, and \texttt{open oven with lantern}; the recorded directional attempts fail.} \\
\multicolumn{3}{@{}p{\textwidth}@{}}{\textbf{GT.} ``Opening the oven reveals a loaf of bread.''} \\
Prediction & ``You can't see any oven here!'' & ``Opening the oven reveals a loaf of bread.'' \\
BLEU & $\approx 0.000$ & $1.000$ \\
\multicolumn{3}{@{}p{\textwidth}@{}}{\textbf{Interpretation.} obs-direct denies the oven's availability; latent correctly predicts the interaction. Both output the correct score tag, $20/400$. The outcome is consistent with state retention, but does not establish which history entries the model used.} \\
\midrule
\multicolumn{3}{@{}p{\textwidth}@{}}{\textbf{J2: Preserving object accessibility.} \texttt{balances\_step12\_k15\_s202}; $t=12$, $k=15$.} \\
\multicolumn{3}{@{}p{\textwidth}@{}}{\textbf{Context.} The earlier \texttt{rezrov box} action opens the cedarwood box and reveals Helistar's grimoire ($o_3$); the grimoire is examined at $o_5$. After a 15-action forecast window, the final action is \texttt{take grimoire}.} \\
\multicolumn{3}{@{}p{\textwidth}@{}}{\textbf{GT.} ``Taken.''} \\
Prediction & ``You can't see any such thing.'' & ``Taken.'' \\
BLEU & $\approx 0.000$ & $1.000$ \\
\multicolumn{3}{@{}p{\textwidth}@{}}{\textbf{Interpretation.} Latent correctly predicts that the grimoire remains accessible, whereas obs-direct reports it as unavailable. Both score tags are correctly $10/51$.} \\
\midrule
\multicolumn{3}{@{}p{\textwidth}@{}}{\textbf{J3: Updating the navigation endpoint.} \texttt{ludicorp\_step37\_k20\_s202}; $t=37$, $k=20$.} \\
\multicolumn{3}{@{}p{\textwidth}@{}}{\textbf{Context.} The window includes movement around the carpark and bike shed, ending with \texttt{east}, \texttt{e}, \texttt{north}. The penultimate true location is Front Of The Office; the last movement reaches Main Entrance.} \\
\multicolumn{3}{@{}p{\textwidth}@{}}{\textbf{GT (excerpt).} ``Main Entrance\ldots The main entrance to the building is plush and opulent. And locked.\ldots''} \\
Prediction & ``Front Of The Office'' & ``Main Entrance\ldots The main entrance to the building is plush and opulent. And locked.\ldots'' \\
BLEU & $\approx 0.000$ & $1.000$ \\
\multicolumn{3}{@{}p{\textwidth}@{}}{\textbf{Interpretation and limit.} Latent predicts the correct final room description; obs-direct returns the preceding location. However, latent's score tag is $15/150$, versus GT $10/150$ and obs-direct $10/150$: this is an observation-prediction win, not complete state accuracy.} \\
\bottomrule
\end{tabularx}
\end{table*}

\begin{table*}[t]
\centering
\small
\setlength{\tabcolsep}{5pt}
\renewcommand{\arraystretch}{1.12}
\caption{Three paired ScienceWorld examples. obs-direct is SFT; latent is \texttt{latent-SR}. Fact-F1 values are the saved scores of full outputs, not scores recomputed on excerpts. Future true observations are used only for retrospective case inspection.}
\label{tab:scienceworld_cases}
\begin{tabularx}{\textwidth}{@{}p{0.13\textwidth}>{\raggedright\arraybackslash}X>{\raggedright\arraybackslash}X@{}}
\toprule
 & \textbf{Raw-history} & \textbf{Latent-based} \\
\midrule
\multicolumn{3}{@{}p{\textwidth}@{}}{\textbf{S1: Tracking a multi-room route.} \texttt{d10a1a445c15\_2\_h8}; $t=2$, $k=8$.} \\
\multicolumn{3}{@{}p{\textwidth}@{}}{\textbf{Context.} The observed prefix ends in the kitchen. Future actions move to the hallway and greenhouse, pick up a shovel and soil, teleport to the hallway, and traverse doors to the greenhouse and then the hallway.} \\
\multicolumn{3}{@{}p{\textwidth}@{}}{\textbf{GT.} ``You move through the door to the hallway.''} \\
Prediction & ``You move to the door between kitchen and hallway.'' & ``You move through the door to the hallway.'' \\
Fact-F1 & $0.000$ & $1.000$ \\
\multicolumn{3}{@{}p{\textwidth}@{}}{\textbf{Interpretation.} obs-direct still references the kitchen--hallway door and does not correctly describe the final traversal; latent predicts the correct destination.} \\
\midrule
\multicolumn{3}{@{}p{\textwidth}@{}}{\textbf{S2: Replacing an outdated room description.} \texttt{c5b6f05af206\_4\_h8}; $t=4$, $k=8$.} \\
\multicolumn{3}{@{}p{\textwidth}@{}}{\textbf{Context.} The last historical observation is ``You move to the kitchen.'' The future window contains \texttt{go to hallway}; the final action is \texttt{look around}.} \\
\multicolumn{3}{@{}p{\textwidth}@{}}{\textbf{GT (excerpt).} ``This room is called the hallway. In it, you see:\ldots a drawing\ldots''} \\
Prediction & ``This room is called the kitchen. In it, you see:\ldots a chair\ldots a counter\ldots'' & ``This room is called the hallway. In it, you see:\ldots a drawing\ldots'' \\
Fact-F1 & $0.000$ & $1.000$ \\
\multicolumn{3}{@{}p{\textwidth}@{}}{\textbf{Interpretation.} obs-direct generates the outdated kitchen scene; latent updates both the room identity and the associated observation.} \\
\midrule
\multicolumn{3}{@{}p{\textwidth}@{}}{\textbf{S3: Predicting the final interaction after relocation.} \texttt{3b6e93f8dd8d\_9\_h5}; $t=9$, $k=5$.} \\
\multicolumn{3}{@{}p{\textwidth}@{}}{\textbf{Context.} From the hallway, execute \texttt{go to kitchen}, \texttt{look around}, \texttt{move apple to blue box}, \texttt{pour apple into metal pot}, and \texttt{open drawer in counter}. The pour action fails, but the final drawer-opening action succeeds.} \\
\multicolumn{3}{@{}p{\textwidth}@{}}{\textbf{GT.} ``The drawer is now open.''} \\
Prediction & ``No known action matches that input.'' & ``The drawer is now open.'' \\
Fact-F1 & $0.000$ & $1.000$ \\
\multicolumn{3}{@{}p{\textwidth}@{}}{\textbf{Interpretation.} obs-direct incorrectly reports failure for the final action; latent predicts the successful state change. obs-direct's output matches the preceding true failure message, but this does not prove that obs-direct internally copied that message.} \\
\bottomrule
\end{tabularx}
\end{table*}

\begin{table*}[t]
\centering
\small
\setlength{\tabcolsep}{5pt}
\renewcommand{\arraystretch}{1.1}
\caption{Three paired CEO examples, showing all four scored fields. obs-direct is SFT; latent is \texttt{latent-SR}. The score is $1-\frac14\sum_f\min\{|\hat y_f-y_f|/\max(100,|y_f|),1\}$; higher is better. Future inflows and cancellations below are ground-truth audit quantities, not assumed model inputs.}
\label{tab:ceo_cases}
\begin{tabularx}{\textwidth}{@{}>{\raggedright\arraybackslash}Xrrr@{}}
\toprule
\textbf{Field} & \textbf{Ground truth} & \textbf{Raw-history} & \textbf{Latent-based} \\
\midrule
\multicolumn{4}{@{}p{\textwidth}@{}}{\textbf{C1: Subscriber depletion.} Session \texttt{33f8310faa3b}; $t=33$, $k=5$. Subscriber stock evolves from $1{,}574$ to $3$: $1{,}574+0-1{,}571=3$.} \\
Cash & $5{,}155{,}166$ & $5{,}165{,}329$ & $5{,}136{,}500$ \\
Individual subscribers & $3$ & $1{,}574$ & $0$ \\
Enterprise seats & $0$ & $0$ & $0$ \\
Open issues & $0$ & $238$ & $0$ \\
Four-field score & --- & $0.500$ & $0.999$ \\
\multicolumn{4}{@{}p{\textwidth}@{}}{\textbf{Interpretation.} obs-direct copies all four initial fields. Latent predicts near-depletion, although $0$ is not the exact subscriber target $3$. Its unscored cancellation forecast remains $509$, versus GT $16$, so this is not evidence of accurate flow prediction throughout the window.} \\
\midrule
\multicolumn{4}{@{}p{\textwidth}@{}}{\textbf{C2: Sustained net outflow.} Session \texttt{7e71cd061842}; $t=27$, $k=9$. The subscriber stock falls from $10{,}206$ to $2{,}131$: $10{,}206+1{,}269-9{,}344=2{,}131$.} \\
Cash & $4{,}499{,}106$ & $4{,}422{,}222$ & $5{,}343{,}825$ \\
Individual subscribers & $2{,}131$ & $10{,}206$ & $3{,}216$ \\
Enterprise seats & $0$ & $0$ & $0$ \\
Open issues & $261$ & $1{,}318$ & $160$ \\
Four-field score & --- & $0.598$ & $0.916$ \\
\multicolumn{4}{@{}p{\textwidth}@{}}{\textbf{Interpretation.} obs-direct retains the initial subscriber and issue stocks. Latent better captures their decline, but its cash forecast is less accurate than obs-direct's; the advantage is in the aggregate score, not every field.} \\
\midrule
\multicolumn{4}{@{}p{\textwidth}@{}}{\textbf{C3: Net growth despite late cancellations.} Session \texttt{5b62a73d6aec}; $t=22$, $k=7$. Subscribers rise from $297{,}368$ to $387{,}099$: $297{,}368+286{,}491-196{,}760=387{,}099$, despite a decline in the final two weeks.} \\
Cash & $3{,}354{,}137$ & $10{,}000{,}000$ & $1{,}930{,}000$ \\
Individual subscribers & $387{,}099$ & $0$ & $397{,}368$ \\
Enterprise seats & $0$ & $0$ & $0$ \\
Open issues & $56{,}235$ & $0$ & $34{,}700$ \\
Four-field score & --- & $0.250$ & $0.656$ \\
\multicolumn{4}{@{}p{\textwidth}@{}}{\textbf{Interpretation.} Latent preserves the direction and approximate scale of accumulated subscriber growth; obs-direct predicts an unsupported collapse. The latent forecast still underestimates cash and open issues.} \\
\bottomrule
\end{tabularx}
\end{table*}

\end{document}